\documentclass[journal]{IEEEtran}
\usepackage[T1]{fontenc}% optional T1 font encoding

\usepackage{cite}
\usepackage{bbm}
\usepackage{amsmath}
\usepackage{amssymb}
\usepackage{amsthm}
\usepackage{dsfont}
\usepackage{graphicx}
\usepackage{multicol}
\usepackage{textcomp}
\usepackage{xcolor}
\usepackage{xspace}
\usepackage{algorithm,algorithmic}
\usepackage{hyperref}
\usepackage{tabularray}
\usepackage{booktabs}
\usepackage{wrapfig}
\usepackage{booktabs}
\usepackage{makecell}
\usepackage{multirow}
\usepackage{booktabs}
\usepackage{caption}
\usepackage{MnSymbol,bbding,pifont}
\newcommand{\cmark}{\ding{51}}%
\newcommand{\xmark}{\ding{55}}%
\DeclareMathOperator*{\sign}{sign}
\usepackage{color,soul}
\definecolor{lightgray}{gray}{.9}

\usepackage{etoolbox}
\usepackage[normalem]{ulem}
\usepackage{caption}
\usepackage{subcaption}
\DeclareMathAlphabet{\mathcal}{OMS}{cmsy}{m}{n}
\makeatletter
\patchcmd{\@makecaption}
{\scshape}
{}
{}
{}
\makeatother

\usepackage[cmintegrals]{newtxmath}

\makeatletter
\newcommand\fs@betterruled{%
  \def\@fs@cfont{\bfseries}\let\@fs@capt\floatc@ruled
  \def\@fs@pre{\vspace*{5pt}\hrule height.8pt depth0pt \kern2pt}%
  \def\@fs@post{\kern2pt\hrule\relax}%
  \def\@fs@mid{\kern2pt\hrule\kern2pt}%
  \let\@fs@iftopcapt\iftrue}
\floatstyle{betterruled}
\restylefloat{algorithm}
\makeatother

\usepackage{fancyhdr}
 
\begin{document}

\title{Federated Hybrid Training and Self-Adversarial Distillation: Towards Robust Edge Networks}

\author{Yu Qiao,~\IEEEmembership{Student Member, IEEE},
        Apurba Adhikary,~\IEEEmembership{Student Member, IEEE},
        Kitae Kim, \\
        Eui-Nam Huh,~\IEEEmembership{Member, IEEE},
        Zhu Han,~\IEEEmembership{Fellow, IEEE},
        and Choong Seon Hong,~\IEEEmembership{Fellow, IEEE}% <-this % stops a space
\IEEEcompsocitemizethanks{

\IEEEcompsocthanksitem Yu Qiao is with the Department of Artificial Intelligence, School of Computing, Kyung Hee University, Yongin-si 17104, Republic of Korea (email: qiaoyu@khu.ac.kr).
\IEEEcompsocthanksitem Zhu Han is with the Department of Electrical and Computer Engineering, University of Houston, Houston, TX 77004 USA, and also with the Department of Computer Science and Engineering, Kyung Hee University, Seoul 446-701, South Korea (e-mail: hanzhu22@gmail.com)
\IEEEcompsocthanksitem Apurba Adhikary, Kitae Kim, Eui-Nam Huh, and Choong Seon Hong are with the Department of Computer Science and Engineering, School of Computing,
Kyung Hee University, Yongin-si 17104, Republic of Korea (e-mail: apurba@khu.ac.kr; glideslope@khu.ac.kr; johnhuh@khu.ac.kr; cshong@khu.ac.kr).
(Corresponding author: Choong Seon Hong).}}

\maketitle

\begin{abstract}
Federated learning (FL) is a distributed training technology that enhances data privacy in mobile edge networks by allowing data owners to collaborate without transmitting raw data to the edge server. However, data heterogeneity and adversarial attacks pose challenges to develop an unbiased and robust global model for edge deployment. To address this, we propose Federated hyBrid Adversarial training and self-adversarial disTillation (FedBAT), a new framework designed to improve both robustness and generalization of the global model. FedBAT seamlessly integrates hybrid adversarial training and self-adversarial distillation into the conventional FL framework from data augmentation and feature distillation perspectives. From a data augmentation perspective, we propose hybrid adversarial training to defend against adversarial attacks by balancing accuracy and robustness through a weighted combination of standard and adversarial training. From a feature distillation perspective, we introduce a novel augmentation-invariant adversarial distillation method that aligns local adversarial features of augmented images with their corresponding unbiased global clean features. This alignment can effectively mitigate bias from data heterogeneity while enhancing both the robustness and generalization of the global model. Extensive experimental results across multiple datasets demonstrate that FedBAT yields comparable or superior performance gains in improving robustness while maintaining accuracy compared to several baselines.
\end{abstract}


\IEEEpeerreviewmaketitle

\section{Introduction}
\IEEEPARstart{S}{ince} significant advancements have been made in the performance of mobile smart devices, such as smartphones, tablets, and IoT sensors, many computing tasks can now be efficiently executed at the edge~\cite{deng2020edge,hua2023edge}. This shift towards edge computing transforms traditional centralized data processing by emphasizing decentralized operations and real-time data analytics to address bandwidth limitations and communication delays~\cite{wu2022node,shang2021nvm}. As a result, multi-access edge computing (MEC) has emerged as the next generation of computing networks, offering reduced latency and improved scalability to meet the demands of modern applications~\cite{wang2023wireless,qiao2023mp}. Meanwhile, as edge devices generate increasingly large amounts of data, MEC also provides a platform for intelligent inference at the edge, supporting applications such as autonomous driving, smart cities, and smart homes~\cite{wang2020convergence}. However, achieving intelligent inference typically requires training a high-performance model, which traditionally involves aggregating large amounts of data from various sources for model training~\cite{qi2023model}.

Recently, this approach has been increasingly replaced by a new paradigm known as federated learning (FL), which addresses issues related to data security, privacy, and communication efficiency~\cite{mcmahan2017communication}. By collaboratively training a global model without sharing local data with a central server or other devices, the federated training paradigm significantly enhances data privacy and reduces the risk of data breaches, positioning it as a promising technology for achieving intelligent inference in edge computing. However, a significant challenge in FL is the non-independent and identically distributed (non-IID) data challenge~\cite{mcmahan2017communication,li2021fedbn}. This challenge arises because the data on each edge device may be influenced by its unique environment, behavior, or other factors, leading to data distributions tailored to its specific context. Consequently, each client’s local model, trained on this non-IID data, may not align with the global model's update direction, potentially causing biases in the global model during inference~\cite{wang2020convergence}.

Moreover, recent studies have shown that similar to centralized model training, federated models are also vulnerable to adversarial attacks, which pose a significant threat to the security and reliability of deployments in real-world scenarios~\cite{zizzo2020fat,hong2023federated,lyu2022privacy}. Specifically, during the model inference stage, attackers can introduce carefully crafted small perturbations (usually imperceptible to the human vision) into test samples to deceive the global model into making incorrect predictions~\cite{qiao2024logit}. Such attacks can lead to serious consequences, including compromising data privacy and degrading model performance. For instance, in the healthcare domain, attackers can cause the model to generate erroneous responses, potentially exposing sensitive patient information~\cite{newaz2021survey}. Similarly, in e-commerce, these attacks can lead to inaccurate product recommendations and the leakage of customers' personal and financial data, negatively impacting the user experience and trust~\cite{tsai2024effective}. Therefore, it is crucial to design a robust FL model capable of defending against such attacks to ensure secure and reliable adoption in real-world applications.

To defend against such attacks, a widely recognized strategy in centralized model training is adversarial training (AT)~\cite{goodfellow2014explaining}, which involves training a robust model using adversarial examples (AEs) generated by adding perturbations to clean examples (CEs). This adversarial training strategy is generally formulated as a min-max problem: the inner optimization seeks to generate the worst-case perturbations to deceive the model using projected gradient descent (PGD)~\cite{madry2018towards}, while the outer optimization minimizes the training loss based on these AEs~\cite{madry2018towards}. Recent research has demonstrated that this strategy is also effective for defending against adversarial attacks in the FL domain~\cite{zizzo2020fat,li2023federated,lyu2022privacy}. Specifically, researchers apply the AT strategy to local model updates during each round of global training, thereby gradually enhancing the robustness of the global model. This approach, known as federated adversarial training (FAT), is considered a promising strategy for achieving robust federated models~\cite{zizzo2020fat}. However, a significant challenge in FAT is that improving robust accuracy (RA) on AEs often reduces clean accuracy (CA) on CEs. This trade-off occurs in both IID and non-IID settings, as shown in Table~\ref{tab:tease_example}. For example, under the vanilla AT strategy adopted in FL, termed FedPGD, adversarial robustness (see AA attack~\cite{croce2020reliable} column) increases by 19.04\% on non-IID data, while clean accuracy (see CA column) decreases by 14.78\% compared to FedAvg~\cite{mcmahan2017communication}. Therefore, mitigating the impact of adversarial attacks and non-IID data is crucial to achieving a robust and unbiased global model in FL, as these factors can significantly reduce CA and RA, affecting model deployment in edge networks.

In this paper, we propose a new FAT framework called Federated hyBrid Adversarial training and self-adversarial disTillation (FedBAT), which comprises two key components. First, to defend against adversarial attacks while maintaining relatively high CA, we introduce a FL-based hybrid-AT strategy that treats the AT process as a data augmentation scheme, which integrates both standard and adversarial training methods during each client’s local updates. Specifically, we jointly optimize two training branches: one branch trains on CEs to achieve high CA, while the other trains on AEs to enhance RA. A coefficient is utilized to adjust the balance between these branches, allowing for flexible tuning of the relative importance of CA and RA. Second, to mitigate the non-IID challenges within the AT framework, we propose regularizing each client’s local updates by aligning each local adversarial representation with its corresponding global clean representation. To be more specific, we derive global representations by averaging local representations, which are obtained from input CEs after applying random augmentations across different clients. We then regularize local updates by aligning local adversarial representations with these global representations that share the same semantic labels. Since global representations are generally less biased than local ones~\cite{li2021model,lu2024federated}, this approach may help ensure that local updates stay close to the global optimum. Additionally, it can facilitate learning augmentation-invariant knowledge and promote consistency between adversarial and clean representations, thereby enhancing the robustness and generalization of the global model. The preliminary version of this
work has been published in~\cite{qiaoICC_2024knowledge}, where we introduced a straightforward knowledge distillation-based regularization strategy for adversarial robustness. The major difference between our work and~\cite{qiaoICC_2024knowledge} is the introduction of a hybrid-AT scheme that integrates standard and adversarial training in local updates, aiming to achieve a balanced CA and RA improvement in FL. Moreover, we propose a self-adversarial distillation strategy based on data augmentation invariance, promoting consistency in the representations of augmented images across different clients. In other words, augmented images of semantically similar images from different clients should be close in the representation space. 

The main contributions of this paper are summarized as follows:
\begin{itemize}
\item We first highlight the adverse effects of adversarial attacks on federated models and the limitation of the vanilla defense strategy adopted in federated settings. To address these issues, we propose a new FedBAT framework that incorporates a hybrid-AT strategy and an augmentation-invariant self-adversarial distillation approach, aiming to mitigate both adversarial attacks and non-IID data challenges. 
\item The proposed hybrid-AT strategy is seamlessly integrated into each local training process, with a coefficient balancing accuracy and robustness. Meanwhile, the augmentation-invariant self-adversarial distillation serves as a regularization term, promoting consistency between local and global models while encouraging the learning of augmentation-invariant features. These strategies complement each other, enhancing the global model's robustness and generalization while maintaining relatively high clean accuracy.
\item We experiment with various datasets, including MNIST~\cite{lecun1998gradient}, Fashion-MNIST~\cite{xiao2017fashion}, SVHN~\cite{netzer2011reading}, Office-Amazon~\cite{gong2012geodesic}, and CIFAR-10~\cite{krizhevsky2009learning}, to evaluate the effectiveness of our proposed method in addressing both adversarial attacks and non-IID data challenges. The results demonstrate that our approach achieves comparable or superior performance in terms of both accuracy and robustness when compared to several baselines.
\end{itemize}

The remaining sections of this paper are structured as follows. Section~\ref{sec:related_work} reviews related work, and Section~\ref{sec:sys_model} outlines the system model and preliminaries. The proposed method is detailed in Section~\ref{sec:method}, and Section~\ref{sec:experiments} presents comprehensive experimental results. Finally, Section~\ref{sec:conclusion} concludes the work.

\section{Related Work}
\label{sec:related_work}

\subsection{Federated Learning}
\label{subsec:fl}
The pioneering work of~\cite{mcmahan2017communication} introduced federated training, a novel distributed learning paradigm in which clients collaborate with an edge server by sharing only model parameters to protect data privacy during model training. In this process, each client optimizes its model using local data and then sends the updated parameters to the edge server for aggregation, repeating this until convergence. However, FL encounters two major challenges: system heterogeneity and statistical heterogeneity, both of which complicate the training and optimization process in the FL environment. Several studies have focused on addressing system heterogeneity from various perspectives. Sageflow~\cite{park2021handling} presents a robust FL framework aimed at resolving the issue of stragglers. FedAT~\cite{chai2021fedat} groups edge devices according to their system capabilities using an asynchronous layer. Similarly, CDFed~\cite{qiao2023cdfed} clusters distributed devices based on their hardware capabilities through a logical layer, effectively reducing the impact of hardware differences. Another main line of research focuses on statistical heterogeneity (a.k.a non-IID data), where local update directions may not align well with the global model, potentially slowing convergence and reducing performance~\cite{hu2023federated_sparsified,li2024representative,shi2024analysis}. To address this challenge, many efforts~\cite{karimireddy2020scaffold,li2021model,le2023fedmekt,tan2022fedproto,qiao2023mp,t2020personalized,zhang2024multi} have been made to mitigate this client drift during local updates. From the perspective of regularization, \cite{le2023fedmekt} employs embedding representation alignment to reduce the discrepancy between local and global models. \cite{li2020federated} addresses this issue by imposing penalties on local training to maintain consistency between local models and the global model. \cite{tan2022fedproto} introduces a method for guiding local training through the exchange of prototypes between clients and the server, thereby avoiding privacy concerns without the need to transmit model parameters. From the perspective of personalized FL, \cite{zhang2024multi} proposes a multi-level personalized FL to address non-IID and long-tailed data challenges. \cite{t2020personalized} employs the Moreau envelope function to separate the optimization of personalized and global models, enabling it to update the global model akin to FedAvg~\cite{mcmahan2017communication} while simultaneously optimizing each client's personalized model based on its local data. Additionally, recent research has explored advanced strategies to enhance FL models from feature-sensitive and constraint-free perspectives. For instance, \cite{yang2024fedfed} partitions data into performance-sensitive and performance-robust features, sharing the former globally to mitigate data heterogeneity while keeping the latter local. \cite{wang2024aggregation} eliminates the need for global aggregation by introducing an aggregation-free FL framework that leverages condensed data and soft labels. Meanwhile, \cite{wang2024dfrd} focuses on privacy concerns by proposing a data-free distillation method using a conditional generator, removing the reliance on real privacy-sensitive datasets. However, despite their effectiveness, these approaches do not address the robustness of FL models against adversarial attacks, which remains an open yet under-explored area.

\subsection{Knowledge Distillation }
\label{subsec:know_distill}
Knowledge distillation (KD)~\cite{hinton2015distilling} is a model compression technique that enhances the performance and efficiency of smaller student models by transferring knowledge from more complex teacher models. The pioneering strategy~\cite{hinton2015distilling} utilizes response-based knowledge~\cite{hinton2015distilling,sun2024logit} such as logit output as the information carrier for the distillation process. In addition, researchers have explored alternative types of knowledge, such as intermediate-level guidance, to provide richer supervision from the teacher model. These include feature-based knowledge~\cite{guan2020differentiable,yu2023data} and relation-based knowledge~\cite{passalis2020probabilistic,xie2024pairwise}. Beyond the issue of what knowledge can be distilled, another line of research investigates how distillation is performed, categorizing methods into online, offline, and self-distillation approaches~\cite{wang2021knowledge}. Since local models and the global model in FL typically share the same network architecture, this paper focuses on self-distillation. In FL, KD has become a crucial technique for enhancing model performance and efficiency by effectively transferring knowledge from the global model to local models~\cite{li2024federated}. There are several strategies based on the core concept of KD, primarily categorized into label distillation~\cite{lu2023federated}, feature distillation~\cite{li2021model}, and model parameter distillation~\cite{li2020federated}. For example,~\cite{lu2023federated} proposes masking out majority label predictions from the global model to enhance distillation's focus on preserving minority label knowledge for each client.~\cite{li2021model} suggests using contrastive learning at the model level by leveraging the similarity between model embeddings to guide local training.~\cite{li2020federated} introduces a proximal term to align local model parameters with those of the global model, ensuring that local updates remain close to the global optimum. Moreover, recent studies have introduced adversarial distilltion~\cite{fang2019data,goldblum2020adversarially}, which focuses on enhancing model robustness by incorporating AEs into the distillation process. For example, FatCC~\cite{qiao2024logit} aligns adversarial and clean features between the local and global models. DBFAT~\cite{zhang2023delving} constraints the global and local models' logits, with the local model's logits being derived from AEs. In this paper, we focus on robust FL and explore augmentation-invariant self-distillation techniques to enhance the robustness of the global model while maintaining relatively higher accuracy in the face of adversarial attacks and non-IID data challenges.

\begin{table}[t]
\centering
\caption{Comparison of CA (\%) and RA (\%) on the Cifar-10~\cite{krizhevsky2009learning} dataset under the AA attack~\cite{croce2020reliable} with a perturbation level of 8.0/255~\cite{goodfellow2014explaining}. The non-IID setting uses a Dirichlet~\cite{balakrishnan2014continuous} distribution with Dir(0.5).}
\label{tab:tease_example}
\vspace{-5px}
\resizebox{0.48\textwidth}{!}{
\begin{tabular}{c|cccc|ccccc}
\toprule
Type & \multicolumn{4}{c|}{Non-IID} & \multicolumn{4}{c}{IID} \\
\midrule
Metric & CA & $\Delta$ & AA & $\Delta$  & CA & $\Delta$ & AA & $\Delta$ \\
\midrule
FedAvg & 58.06 & - & 1.20 & - & 59.98 & - & 1.22 & - \\
FedPGD & 43.28 & -14.78 & 20.24 & +19.04 & 43.68 & -16.30 & 19.90 & +18.68 \\
FedBAT & 45.98 & -12.08 & 20.70 & +19.50 & 46.20 & -13.78 & 20.68 & +19.46\\
\bottomrule
\end{tabular}}
\vspace{-5px}
\end{table}

\begin{table}[!t]
\caption{Summary of Notations.}
\label{tab:notations}
\vspace{-0.1cm}
\centering
\begin{tabular}{|c|l|}
\hline
Notation & Description\\
\hline
$\mathcal F_i(\cdot)$ & Shared global model for each client \\
$\theta$ & Model parameters \\
$\mathcal{D}_i$ & Local dataset for each client\\
$P_j(\cdot)$ & Class probability after softmax operation\\
$\boldsymbol {x}_i$ & Local input images for each client\\
$y_i$ & Labels \\
$\mathbb{I}(\cdot)$ & Indicator function \\
$\mathcal{C}$ & Set of classes \\
$D_i$ & Size of the local dataset for each client\\
$\mathcal{L}_i$ & Local objective for each client \\
$\nabla \mathcal{L}_i$ & Gradient for each local objective\\
$\eta$ & Learning rate \\
$N$ & Number of clients\\
$\sign(\cdot)$ & Sign function \\
$\alpha$ & Step size of PGD attack algorithm \\
$\delta$ & Perturbations \\
$\boldsymbol{x}_i^{adv}$ & Adversarial examples of each client \\
$\mathcal{L}_i^{adv}(\cdot)$ & Local AT objective for each client \\
$\mathcal{L}_{adv}(\cdot)$ & Global AT objective \\
$\lambda$ & Control the tradeoff between CA and RA \\
$\mathcal{L}_{FHA}$ & Hybrid-AT loss for each client \\
$f_i^e(\cdot)$ & Feature extraction component \\
$f_i^d(\cdot)$ & Decision-making component \\
$X_{i,j}$ & Local augmented representation of class $j$ for client $i$ \\
$\bar{X}_j$ & Global augmented representation of class $j$ \\
$\bar{X}$ & Set of all global augmented representation \\
$\Vert \cdot \Vert_2^2$ & Squared $\ell_2$ distance \\
$\mathcal{L}_{ASD}$ & Augmentation-invariant self-adversarial distillation loss \\
$\hat{\boldsymbol{x}}_i$ & Augmented input images of each client \\
\hline
\end{tabular}
\vspace{-5px}
\end{table}

\subsection{Adversarial Attack and Defense}
\label{subsec:adv_attack}
Similar to findings in centralized model training, where deep neural networks (DNNs) are known to be vulnerable to AEs~\cite{szegedy2013intriguing}, recent research~\cite{qiao2024noms_fedalc,zizzo2020fat} has demonstrated that FL models are also susceptible to adversarial attacks. The AEs are typically generated by introducing imperceptible perturbations to test images, and these perturbed images are then used to mislead the global model during inference, resulting in inaccurate predictions. Fast gradient sign method (FGSM)~\cite{goodfellow2014explaining} is the first technique developed for generating AEs, while PGD~\cite{madry2018towards} and basic iterative method (BIM)~\cite{kurakin2018adversarial} are iterative extensions of FGSM. Subsequently, several advanced variants have been developed to craft AEs with stronger attack capabilities, including Square~\cite{andriushchenko2020square}, Carlini and Wagner (C\&W)~\cite{carlini2017towards}, and AutoAttack (AA) attacks~\cite{croce2020reliable}. Another line of research focuses on defending against these attack methods, using approaches such as data augmentation~\cite{rebuffi2021data,zhao2020maximum}, regularization~\cite{li2021towards,jakubovitz2018improving}, robust model architectures~\cite{akhtar2021advances,devaguptapu2021adversarial}, and AT~\cite{zhang2019limitations,goodfellow2014explaining}. Among these, AT is widely recognized as the most effective defense strategy~\cite{zhang2019limitations}. Recently, several works~\cite{zizzo2020fat,hong2023federated,chen2022calfat,zhang2023delving} have successfully applied AT strategies in the FL domain to develop robust FL models. For example, ~\cite{zizzo2020fat,hong2023federated} suggests applying AT to a subset of clients while allowing the remaining clients to perform standard training. \cite{qiao2024logit} enhances the global model's robustness via feature contrast between the local model and the global model. \cite{chen2022calfat,zhang2023delving} propose reweighting each client's logit output to improve the robustness of the global model within non-IID data challenges. Orthogonal to the existing works, our paper proposes hybrid-AT and self-adversarial distillation strategies for each local update, aiming to improve the robustness and generalization of the global model against adversarial attacks and non-IID challenges. Table~\ref{tab:notations} provides a summary of the notations used in this paper.

\section{System Model and Preliminaries}
\label{sec:sys_model}
\subsection{Non-Robust Federated Learning Model}
We consider a FL setting with $N$ edge clients and one edge server, where each client $i$ possesses its own private and sensitive dataset $\mathcal{D}_i= {(\boldsymbol {x}_i, y_i)}$ of size $D_i$, with $\boldsymbol {x}_i$ representing the input images and $y_i$ denoting the corresponding labels. To build a more generalizable shared global model that can adapt to each client, clients cooperate with the edge server to train this model. At each global communication round, each client optimizes the shared global model, denoted as $\mathcal F_i(\theta; \boldsymbol {x}_i)$, based on its own dataset. The empirical risk (e.g., cross-entropy (CE) loss) for client $i$ with one-hot encoded labels can then be defined as follows~\cite{mcmahan2017communication}:
\begin{equation} \label{one_hot}
   {f_i} (\theta; \boldsymbol{x}_i, y_i) = -\sum_{j=1}^{C} \mathbb{I}_{y_i = j}\log P_j(\mathcal F_i(\theta; \boldsymbol {x}_i)),
\end{equation}
where $P_j(\mathcal F_i(\theta; \boldsymbol {x}_i))$ represents the probability of the sample $\boldsymbol {x}_i$ being classified as the $j$-th class, $\theta$ denotes the shared model parameters of global model, $\mathbb{I}_{(\cdot)}$ is the indicator function, and $C$ is the number of classes in the label space $\mathcal{C} = \{1,..,C\}$. Therefore, the local objective for each client aims to minimize the local loss $\mathcal{L}_i$, as follows:
\begin{equation} \label{local_loss}
   {\mathcal{L}_i} (\theta; \boldsymbol{x}_i, y_i) = \frac {1}{D_i} \sum_{i \in \mathcal{D}_i} f_i(\theta; \boldsymbol{x}_i, y_i).
\end{equation}
Subsequently, each client performs local stochastic gradient descent (SGD) to optimize its local objective, as follows:
\begin{equation} \label{sgd}
\theta_{t+1}  =  \theta_{t} - \eta \nabla \mathcal{L}_i(\theta_{t}; \boldsymbol{x}_i, y_i),
\end{equation}
where $\nabla \mathcal{L}_i(\theta_{t}; \boldsymbol{x}_i, y_i)$ represents the gradient of the loss for client $i$ at global round $t$, $\theta_{t+1}$ is the updated model parameters at global round $t+1$, and $\eta$ is the learning rate.

Finally, the global objective $\mathcal{L}(\theta)$ is to minimize the aggregate of local losses across these local clients, which can be denoted as follows:
\begin{equation} \label{global_loss}
    \mathop {\min}_{\theta} \mathcal{L}(\theta) = \sum_{i \in \mathcal{N}} \frac {D_i}{\sum_{i \in \mathcal{N}}D_i} {\mathcal{L}_i} (\theta; \boldsymbol{x}_i, y_i),
\end{equation}
where $\mathcal{N}$ denotes the set of distributed devices with $\mathcal{N} = \{1,...,N\}$.

\begin{figure}[t]
\centering
\includegraphics[width=0.49\textwidth]{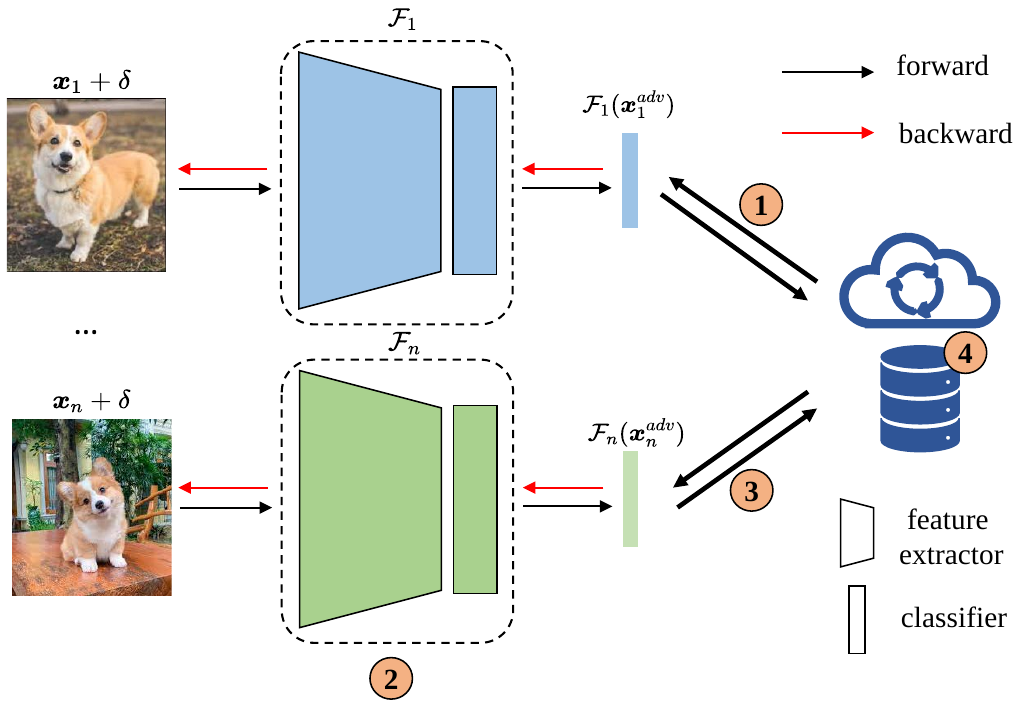}
\caption{Illustration of the vanilla robust FL framework (FedPGD), which serves as the basic structure for robust federated systems. $\mathcal{F}_1$ and $\mathcal{F}_n$ denote the local models of clients $1$ and $n$, respectively. $\mathcal{F}_1(\boldsymbol{x}_1^{adv})$ and $\mathcal{F}_n(\boldsymbol{x}_n^{adv})$ represent the predictions using AEs, with the model updated by minimizing the distance between these predictions and the ground truth labels. FedBAT modifies both the local update process~\textcircled{2} and the global aggregation process~\textcircled{4} to enhance robustness while maintaining accuracy.}
\label{fig:vanilla_robust_FL}
\vspace{-5px}
\end{figure}

\subsection{Robust Federated Learning Model}
\label{sec:robust_FL}
To obtain a robust global model in FL scenarios, each client typically optimizes its local model using AEs instead of CEs during its local update process, while the global model is updated in the same way as in traditional federated training~\cite{zizzo2020fat}. For any given client, the AEs are usually generated using the PGD algorithm, which can be updated as follows:
\begin{equation}
\label{eq:pgd}
\boldsymbol x_i^{t+1}=\Pi_{\boldsymbol{x_i} + \delta}\left(\boldsymbol x_i^t+\alpha\sign(\nabla_{\boldsymbol x^t}  \mathcal{L}_i(\theta; \boldsymbol{x}_i^t, y_i) \right),
\end{equation}
where $\boldsymbol x_i^{t+1}$ denotes the generated AE at $t+1$ step, $\Pi_{\boldsymbol x_i + \delta}$ represents the projection function that projects the $\boldsymbol x_i$ onto the $\epsilon$-ball centered at $\boldsymbol x_i^0$, $\alpha$ represents the step size, and $\sign(\cdot)$ denotes the sign function. To ensure that the perturbation $\delta$ remains imperceptible to human vision, it is typically constrained by an upper bound  $\epsilon$ on the $\ell_\infty$-norm (i.e., $||\delta||_\infty \leq \epsilon$). Once the PGD algorithm has completed, the AEs for each client can be defined as follows:
\begin{equation} \label{AEs_gen}
   \boldsymbol{x}_i^{adv} = \boldsymbol{x}_i + \delta.
\end{equation}

Subsequently, following the practice of AT commonly used in centralized training~\cite{athalye2018robustness}, the generated AEs are incorporated in the local update process of each client's model, as formulated below:
\begin{equation} \label{AT_Local}
   \mathop {\min}_{\theta} \mathbb E_{(\boldsymbol{x}_i, y_i) \sim \mathcal{D}_i} \left[\mathcal{L}_i^{adv}(\theta; \boldsymbol{x}_i^{adv}, y_i)\right],
\end{equation}
where $\mathcal{L}_i^{adv}(\theta; \boldsymbol{x}_i^{adv}, y_i)$ denotes the local AT objective for each client, and the objective of (\ref{AT_Local}) is to optimize the model parameters so that the model remains accurate in classifying samples even after perturbations (i.e., the AEs generated by (\ref{AEs_gen})), thereby enhancing each local model's resilience to adversarial attacks.

Finally, after each local client completes its local AT based on (\ref{AT_Local}), the client uploads its adversarially trained model parameters to the server for aggregation. The aggregated parameters are then distributed back to the clients to begin the next global round. This process continues until convergence or the specified number of global communication rounds is reached. Therefore, (\ref{global_loss}) can be reformulated as the robust optimization process for the global objective, outlined as follows:
\begin{equation} \label{global_loss_at}
    \mathop {\min}_{\theta} \mathcal{L}_{adv}(\theta) = \sum_{i \in \mathcal{N}} \frac {D_i}{\sum_{i \in \mathcal{N}}D_i} {\mathcal{L}_i^{adv}} (\theta; \boldsymbol{x}_i^{adv}, y_i),
\end{equation}
where $\mathcal{L}_{adv}(\theta)$ represents the global AT loss. This formulation applies AT to local clients based on their own data to enhance their robustness, which in turn contributes to the overall improvement of the global model's robustness after federated training processes. In this paper, we follow this training paradigm but focus on refining the local AT process by employing a hybrid-AT strategy and self-distillation techniques to address both adversarial attacks and non-IID data challenges.

% \vspace{-5px}
\section{Methodology}
In this section, we first present the vanilla robust federated learning framework. Next, we introduce the proposed FedBAFT framework, followed by an in-depth explanation of FL-based hybrid adversarial training and augmentation-invariant self-adversarial distillation. Finally, we provide the overall objective of FedBAFT.
\label{sec:method}
\subsection{Vanilla Robust FL Framework}
The vanilla robust FL framework, depicted in Figure~\ref{fig:vanilla_robust_FL}, provides the basic structure for robust federated systems and involves four main steps. For clarity, this framework is referred to as FedPGD in this paper. First, the server distributes the initialized global model to the clients in~\textcircled{1}. Next, each client updates the model parameters using adversarial training (AT) based on its local dataset via (\ref{AT_Local}) in~\textcircled{2}. Subsequently, all clients send their updated model parameters back to the server for aggregation in~\textcircled{3}. Finally, the server aggregates the received model parameters via (\ref{global_loss_at}) in~\textcircled{4}, and this process is repeated until convergence. FedBAT enhances this framework by modifying both the local update process in~\textcircled{2} and the global aggregation process in~\textcircled{4} through the introduction of hybrid-AT schemes and self-distillation strategies to improve robustness while maintaining accuracy in federated systems. The following sections first introduce our proposed robust FedBAT framework, and then detail the design of the hybrid-AT and self-distillation strategies.

\begin{figure}[t]
\centering
\includegraphics[width=0.49\textwidth]{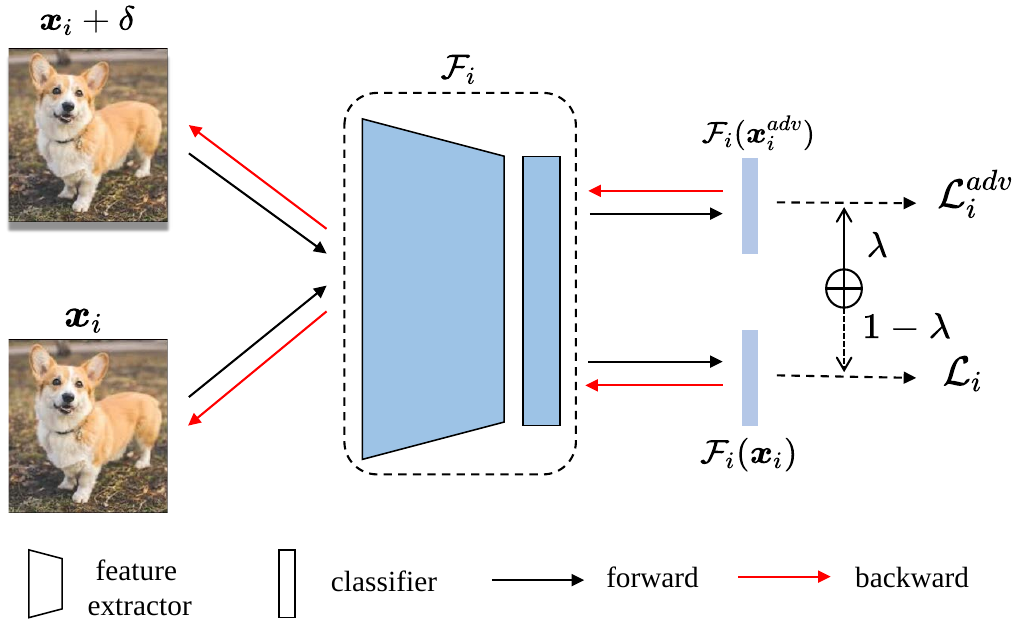}
\caption{Illustration of the proposed FL-based hybrid-AT strategy designed to balance the CA and RA. $\mathcal{F}_i$ denotes the local model of an arbitrary client. $\mathcal{F}_i(\boldsymbol{x}_i^{adv})$ and $\mathcal{F}_i(\boldsymbol{x}_i)$ represent the model's predictions using AEs and CEs, respectively. A coefficient $\alpha$ is used to balance the trade-off between $\mathcal{L}_i^{adv}$ (CE loss) and $\mathcal{L}_i$ (CE loss) during the local training.}
\label{fig:hybrid_at}
\vspace{-3px}
\end{figure}

\begin{figure*}[t]
\centering
\includegraphics[width=0.98\textwidth]{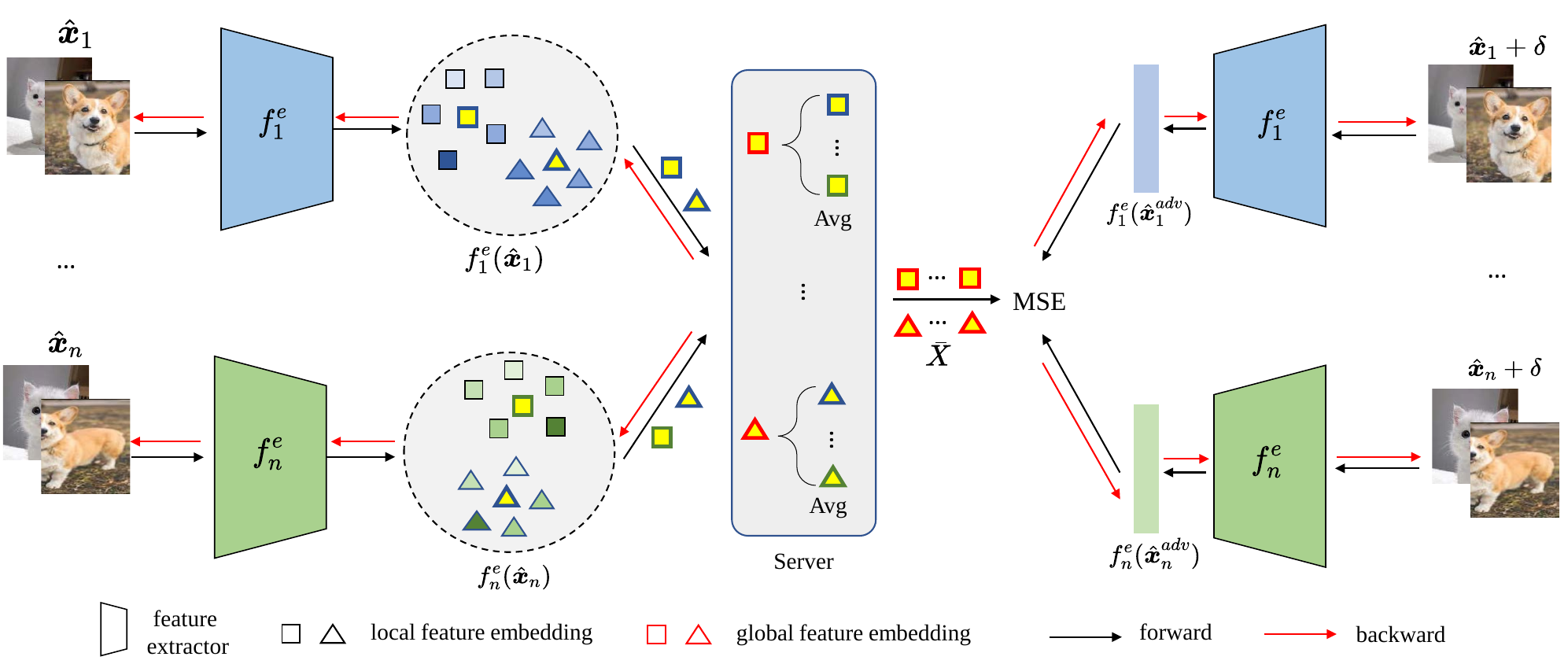}
\caption{Illustration of the proposed self-adversarial distillation strategy designed to address the non-IID challenge. For an arbitrary client, $\hat{\boldsymbol{x}}_i$ represents the augmented CEs, while $\hat{\boldsymbol{x}}_i + \delta$ denotes the augmented AEs. $f_i^{e}$ denotes the feature extractor, and $f_i(\hat{\boldsymbol{x}}_i^{adv})$ and $f(\hat{\boldsymbol{x}}_i)$ represent the model's feature embeddings obtained from AEs and CEs, respectively. $X$ denotes the averaged global clean feature embeddings. An MSE loss function is introduced to align the global clean embeddings with their semantically same local adversarial embeddings.}
\label{fig:self_distillation}
\end{figure*}

\vspace{-0.1cm}

\subsection{Proposed FedBAT Framework}
Different from the standard robust FL framework, the proposed FedBAT framework incorporates two carefully designed components: FL-based hybrid-AT and augmentation-invariant self-adversarial distillation. The former is illustrated in Figure~\ref{fig:hybrid_at}, where we propose a hybrid-AT framework with two branches: one branch is dedicated to CA and the other to RA. Both branches are trained using CE loss. A coefficient $\alpha$ is introduced to balance these two branches during training. Moreover, since global embeddings are generally less biased than local ones, we propose a self-distillation strategy that transfers knowledge from the global model to the local models, as illustrated in Figure~\ref{fig:self_distillation}. An MSE loss is introduced to align the local embeddings with their corresponding global embeddings, ensuring that the local model remains close to the global optimum. The following sections detail the design components of our hybrid-AT and self-distillation strategies.

\subsection{FL-based Hybrid-AT}
\label{subsec:local_hybrid_at}
Following previous robust FL design~\cite{zizzo2020fat}, we find that applying AT locally can enhance the global model's robustness against adversarial attacks, but it often leads to a decline in performance on CEs (unperturbed samples), as illustrated in Table~\ref{tab:tease_example}. Again, we observe that the table reveals a notable drop in accuracy on CEs (i.e., the CA column in the table). For example, the accuracy of FedPGD decreases by 14.78\% and 16.30\% in non-IID and IID settings, compared to vanilla FedAvg. To mitigate this performance decline, we propose using AT as a data augmentation strategy called FHA (FL-based hybrid-AT). This approach involves two training branches that are jointly optimized for both CEs and AEs. For clarity, we consider an arbitrary client within the federated learning framework and illustrate its local training process based on the proposed hybrid-AT approach in Figure~\ref{fig:hybrid_at}. This training scheme aims to minimize the following objective:
\begin{equation} \label{fl_hybrid_at}
   \mathcal{L}_{FHA} = (1 - \lambda) \mathcal{L}_i(\theta; \boldsymbol{x}_i, y_i) + \lambda \mathcal{L}_i^{adv}(\theta; \boldsymbol{x}_i^{adv}, y_i),
\end{equation}
where $\lambda$ controls the tradeoff between CA and RA. During local training, this method enables each client to update the model by simultaneously learning from both CEs and AEs via SGD as described in (\ref{sgd}).

However, given that non-IID data is prevalent in FL environments, even though each client might achieve a balance between robustness and accuracy using (\ref{fl_hybrid_at}), the local update directions may diverge from the global updates, similar to traditional FL approaches\cite{zizzo2020fat,mcmahan2017communication}. This divergence may be due to the local training process potentially overfitting to the client's own data, which in turn may further hinder the global model's generalization ability. To address this issue, the next section introduces a self-distillation strategy that encourages each client to align its local representations with the global representation, thereby reducing divergence between local and global update directions and consequently mitigating the challenges posed by non-IID data.
    
\subsection{Augmentation-Invariant Self-Adversarial Distillation}
\label{subsec:self_distill}
Consider an arbitrary client with a local model that can be generally divided into two main components: the feature extraction component $f_i^e(\cdot)$ and the decision-making component $f_i^d(\cdot)$. The feature extraction component is responsible for extracting features from the input data, while the decision-making component leverages these features to make final predictions. By comparing the prediction with the ground truth labels, each local model is updated iteratively to minimize the loss and thus improve its accuracy on its own local dataset. However, since the data distribution is non-IID across clients, each client has its own unique data distribution. This can cause local models to become highly specialized to their specific data, resulting in updates that may push the global model in conflicting directions~\cite{qiao2024fedccl,zhao2023ensemble,chen2023fraug}. Consequently, this misalignment might weaken the global model's generalization ability. To address this misalignment, we propose a novel approach called augmentation-based local and global feature alignment in the FAT environment. The calculated features are privacy-friendly as they are represented by one-dimensional vectors and are obtained through multiple averaging operations. Since both the local and global models share the same architecture, we refer to this method as augmentation-based self-distillation (ASD).

\textbf{Local Augmented Representation.} For an arbitrary client, to derive the local augmented representation for each class, we begin by applying random augmentation such as cropping, flip, and rotation to obtain the augmented local sample $\{\hat{\boldsymbol{x}}_i, y_i\}$ during training in each mini-batch. These augmented samples are then encoded into $Z_i = f_i^e(\hat{\boldsymbol{x}}_i)$ using the feature extraction component $f_i^e(\cdot)$. To represent each semantic label, we compute the averaged representation of all augmented images corresponding to that label, as follows:
\begin{equation} \label{local_feature}
\begin{aligned}
 {X}_{i,j} &=\frac{1}{S_{i,j}} \sum_{S_{i,j}} Z_{i,j}, \quad i \in \mathcal{N}, \quad j \in \mathcal{C}, \\
 {X} &= [{X}_{i,1},..., {X}_{i,j},..., {X}_{i,C}],
 \end{aligned}
\end{equation}
where $X_{i,j}$ denotes the local representation of the $j$-th class for client $i$, ${X}$ represents the local representation set of client $i$, and $S_{i,j}$ indicates the size of class $j$ for client $i$. The key idea is that, despite different augmentations of the same semantic label resulting in varied images, their representations should remain close to each other and not deviate significantly from their mean value.

\begin{figure}[t]
\centering
\includegraphics[width=0.49\textwidth]{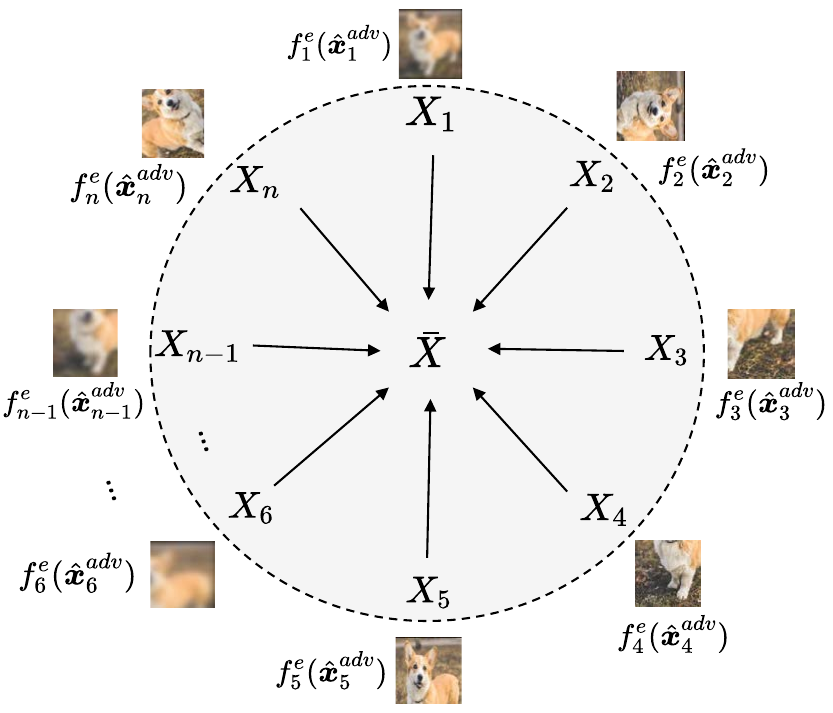}
\caption{Illustration of the key idea behind the proposed self-adversarial distillation. Different augmentations are employed for the same semantic image across various clients, and the resulting representations should be invariant. The goal is to minimize the distance between the representations of the augmented AEs from each client and the mean representation of the corresponding semantically identical CEs.}
\label{fig:aug_align}
\end{figure}

\textbf{Global Augmented Representation.} Motivated by the findings in~\cite{li2021model,tan2022fedproto,mu2023fedproc}, which reveal that the shared global representation presents less bias than local representations in a non-IID data environment, we transmit the local augmented representations calculated in (\ref{local_feature}) to the edge server for aggregation. An averaging operation is then performed at the edge server to obtain the global augmented representation, as detailed below:
\begin{equation} \label{global_feature}
\begin{aligned}
\bar{X}_j &= \frac{1}{N} \sum_{i\in \mathcal{N}} X_{i,j} , \quad i \in \mathcal{N}, \quad j \in \mathcal{C}, \\
\bar{X} &= [\bar{X}_1,..., \bar{X}_j,..., \bar{X}_C],
\end{aligned}
\end{equation}
where $\bar{X}_j$ denotes the global augmented feature of the $j$-th class, and $\bar{X}$ denotes the set of all global augmented features. The key idea, similar to the process for calculating local augmented features, is that the representations of images with the same semantic label from different clients should not deviate significantly from their mean value. Additionally, since these augmented features are averaged across distributed clients and thus can be considered less biased compared to local ones, the global augmented representations can be regarded as pseudo-ground truth labels to guide each local distillation learning process.

\textbf{Augmentation-Invariant Self-Adversarial Distillation.} To achieve alignment between local and global representations, we encourage each local augmented adversarial representation to match the global augmented clean representation, as illustrated in Figure~\ref{fig:self_distillation}. During each global communication round, for a given client, the encoded local adversarial representation $f_i^e(\hat{\boldsymbol{x}}_i)$ is obtained using the feature extraction component $f_i^e$, with $\hat{\boldsymbol{x}}_i + \delta$ as input. The local adversarial representations are then encouraged to align with the global clean representations, as derived using (\ref{global_feature}), by minimizing the mean square error (MSE) loss. Therefore, the ASD loss can be defined as follows:
\begin{equation} 
\label{eq:mse_loss}
\mathcal{L}_{ASD} = \Vert(f^{e}_{i}(\hat{\boldsymbol {x}}_i^{adv})- \bar{X})\|_2^2,
% \vspace{-15px}
\end{equation}
where $\Vert \cdot \Vert_2^2$ denotes the squared $\ell_2$ distance used to measure the difference between the local adversarial feature and the global clean feature. 

We further illustrate the basic idea of the augmentation-invariant adversarial self-distillation scheme in Figure~\ref{fig:aug_align}. It demonstrates that representations of different augmentations from various clients sharing the same semantic label should remain invariant. In other words, images with the same semantic label should be close to each other in the representation space. Further, by aligning local adversarial representations with global clean representations, the model is supposed to learn more consistent features and offer several benefits. First, by leveraging adversarial-clean representation pairing, the model is expected to enhance its adversarial robustness by making the adversarial features consistent with the clean discriminative features. Second, since global representations are less biased compared to local ones, this approach is also expected to mitigate the challenges posed by non-IID data.

\subsection{Overall Objective}
\label{subsec:overall_obj}
Our proposed FedBAT framework comprises two main components. First, to enhance robustness against adversarial attacks while maintaining relatively high accuracy, we introduce a FL-based hybrid-AT scheme. This scheme allows for flexible adjustment of the balance between accuracy and robustness. Second, to mitigate the challenges posed by non-IID data, we introduce an augmentation-invariant self-adversarial distillation strategy. This approach aligns local adversarial representations with their corresponding global clean counterparts, facilitating the learning of augmentation-invariant knowledge and promoting consistency between adversarial and clean representations. As a result, it enhances the robustness and generalization capabilities of the global model. Finally, the overall objective in (\ref{global_loss_at}) can be reformulated as follows:
\begin{subequations}
\begin{align}
\label{overall_at1}
   \mathcal{L}_i^{adv} &= \mathcal{L}_{FHA} + \mathcal{L}_{ASD}, \\
   \label{overall_at2}
   \mathcal{L}_{adv} &= \sum_{i \in \mathcal{N}} \frac {D_i}{\sum_{i \in \mathcal{N}}D_i} \mathcal{L}_i^{adv},
\end{align}
\end{subequations} 
where $\mathcal{L}_i^{adv}$ denote the training objective for each client, and $\mathcal{L}_{adv}$ represents the final overall objective. The detailed training process for FedBAT is outlined in Algorithm~\ref{alg:fedbat}. The algorithm takes heterogeneous datasets and training parameters from different clients as input. Once the federated system is initialized, the proposed FedBAT training process is executed from lines 2 to 10. During each global iteration, all clients perform local training in parallel, as outlined in lines 3 to 6. For each client, random data augmentation is performed in line 15, followed by the calculation of local features in line 17. Subsequently, the FL-based hybrid-AT strategy is implemented in line 19, followed by the self-adversarial distillation strategy in line 21. The aggregation of each client's local objectives is then performed in line 23, followed by SGD-based optimization of the model parameters in line 25. Finally, in line 28, each client transmits its updated model parameters and computed local features back to the server. The server then aggregates the model parameters in line 10 and performs global feature aggregation in line 8, continuing with the next global iteration until all global rounds $T$ are completed.

\begin{algorithm}[t]        \captionsetup{justification=raggedright,singlelinecheck=false}
    \captionsetup[sub]{justification=centering} % C
    \caption{FedBAT} 
    \label{alg:fedbat} 
    \begin{algorithmic}[1]
        \REQUIRE ~~ \\
        Dataset $\mathcal{D}_i$ of each client, $\theta_i$, number of clients $N$.
        \ENSURE ~~ \\
        Robust global model $\theta$.
        \STATE \textbf{Initialize $\theta^0$}.
        \FOR{ $t$ = 1, 2, ..., $T$} 
            \FOR{ $i$ = 0, 1,..., $N$ \textbf{in parallel}}
                \STATE Send global model $\theta^t$ to client \textit{i}
                \STATE {$\theta^t, {X}_{i} \gets \textbf{LocalUpdate}(\theta^t$)}
            \ENDFOR
            \STATE {\textcolor{gray}{\text{/* Global feature calculation */}}}
            \STATE {$\bar{X} \gets \frac{1}{N} \sum_{i\in \mathcal{N}} X_{i,j}$} via (\ref{global_feature})
            \STATE {\textcolor{gray}{\text{/* Overall objective */}}}
            \STATE {$\mathcal{L}_{adv} \gets \sum_{i \in \mathcal{N}}\frac {D_i}{\sum_{i \in \mathcal{N}}D_i} \mathcal{L}_i^{adv}$ by (\ref{overall_at2})}
        \ENDFOR \\
        \textbf {LocalUpdate($\theta^t$, $\bar{X}$)}
        \FOR{ each local epoch }
        \FOR{ each batch ($\boldsymbol{x}_i$; $y_i$) of $\mathcal{D}_i$}
        \STATE {$\boldsymbol{x}_i^{adv} \gets \boldsymbol{x}_i + \delta$} by (\ref{AEs_gen})
        \STATE{ $\hat{\boldsymbol{x}}_i$, $\hat{\boldsymbol{x}}_i^{adv} \overset{\text{augmentation}}{\gets} \boldsymbol{x}_i$, $\boldsymbol{x}_i^{adv}$ }
        \STATE{\textcolor{gray}{\text{/* Local feature calculation */}}}
        \STATE{${X}_{i,j} \gets \frac{1}{S_{i,j}} \sum_{S_{i,j}} Z_{i,j}$} by (\ref{local_feature})\\
        \STATE{\textcolor{gray}{\text{/* FL-based hybrid-AT */}}}
        \STATE{$\mathcal{L}_{FHA} \gets (1-\lambda) \mathcal{L}_i+ \lambda \mathcal{L}_i^{adv}$} via (\ref{fl_hybrid_at})
        \STATE{\textcolor{gray}{\text{/* Self-adversarial distillation */}}}
        \STATE {$\mathcal{L}_{ASD} \gets \|(f^{e}_{i}(\hat{\boldsymbol {x}}_i^{adv}) - \bar{X})\|_2^2$} via (\ref{eq:mse_loss})
        \STATE \textcolor{gray}{\text{/* Local objective for each client */}} 
        \STATE $\mathcal{L}_i^{adv} \gets \mathcal{L}_{FHA} + \mathcal{L}_{ASD}$ by (\ref{overall_at1}) \\
        \STATE \textcolor{gray}{\text{/* Local model parameters optimize */}} 
        \STATE {$ \theta_{t+1}  \gets  \theta_{t} - \eta \nabla \mathcal{L}^{{adv}}_i$ by (\ref{sgd})}
    \ENDFOR
    \ENDFOR
        \RETURN $\theta_i^t$, ${X}_{i}$
    \end{algorithmic}
    % \vspace{-5px}
\end{algorithm}

\subsection{Complexity Analysis} 
We focus on a single global iteration of our proposed algorithm to analyze its computational complexity, as the algorithm comprises multiple similar global iterations. To simplify our analysis, we ignore the data transmission and reception between the global model and the local models and instead concentrate on the main computational complexity during local training, which typically demands more resources and where the key components of our proposal are executed. For our analysis, we consider a fully connected (FC) neural network in which each layer has an equal parameter size, denoted as $M$. Given that the computational complexity of a forward pass can be generally represented as matrix-vector multiplication, and we exclude bias parameters and activation function calculations for simplicity, the time complexity for forward propagation in the FC network during local model training can be expressed as $\mathcal{O}(E \times B \times n_{\text{layer}} \times M^2)$. Here, $B$, $E$, and $n_{\text{layer}}$ represent the batch size, the number of local epochs, and the number of layers in the FC network, respectively. Similarly, for backward propagation, the complexity is generally of the same order as forward propagation. Therefore, the overall time complexity for each local training step is $\mathcal{O}(E \times B \times n_{\text{layer}} \times M^2)$. This complexity applies to each client individually, and since the local updates are performed in parallel by $N$ clients, the overall complexity per global iteration is $\mathcal{O}(E \times B \times n_{\text{layer}} \times M^2)$.

In each global iteration, the server computes the average feature $\bar{X}$ by aggregating the feature vectors $X_i$ from all $N$ clients, with a computational complexity of $\mathcal{O}(N \times C)$. Given that both $N$ and $C$ are typically much smaller than the total complexity of local training, this step is negligible in comparison and does not dominate the overall computational cost. For local feature calculation, each sample within a mini-batch should be averaged, resulting in a time complexity of $\mathcal{O}(E \times B)$. In FL-based hybrid-AT, both clean and adversarial samples are processed within each batch, leading to a time complexity of $\mathcal{O}(2 \times E \times B \times n_{\text{layer}} \times M^2) = \mathcal{O}(E \times B \times n_{\text{layer}} \times M^2)$, where the factor of 2 accounts for handling both sample types. For self-adversarial distillation, the process involves comparing each local feature in a mini-batch with the corresponding global features, resulting in a time complexity of $\mathcal{O}(E \times B \times C)$. In summary, the time complexity for the proposed components is $\mathcal{O}(N \times C + E \times B \times (1 + n_{\text{layer}} \times M^2 + C))$. Since the model parameter $M$ is typically large and dominates the computational time complexity, the time required to compute the proposed component becomes negligible compared to the model updating process.

\begin{table*}[t]
\centering
    \caption{Clean accuracy (\%) and robust accuracy (\%) comparison of FedBAT and other baselines on MNIST and Fashion-MNIST under non-IID setting with Dir(0.5). The best results for federated defense methods are in \textbf{bold}.}
    \label{tab:M_F_CA_RA_comparison}
    \vspace{-0.2cm}
\resizebox{0.99\textwidth}{!}{
\begin{tabular}{l|ccccccc|ccccccc}
\toprule
Dataset & \multicolumn{7}{c|}{MNIST} & \multicolumn{7}{c}{Fashion-MNIST} \\
\midrule
Methods & Clean & FGSM & BIM & PGD-40 & PGD-100 & Square & AA & Clean & FGSM & BIM & PGD-40 & PGD-100 & Square & AA \\
\midrule
FedAvg~\cite{mcmahan2017communication} & 97.58 & 5.78 & 18.02 & 0.50 & 0.16 & 0.06 & 0.00 & 79.18 & 27.62 & 13.02 & 12.50 & 12.30 & 12.00 & 10.34 \\
MixFAT~\cite{zizzo2020fat} & 96.22& 61.52 & 75.12 & 38.04 & 16.32 & 10.24 & 7.18 & 71.22 & 53.04 & 47.42 & 46.96 & 45.16 & 44.10 & 43.72 \\
FedPGD~\cite{madry2018towards} & 96.44 & 63.30 & 74.84 & 41.44 & 21.24 & 14.30 & 11.14 & 68.00 & 54.04 & 50.34 & 49.34 & 49.22 & 45.20 & 45.04\\
FedALP~\cite{kannan2018adversarial} & 93.02 & 63.84 & 73.54 & 46.54 & 30.12 & 15.86 & 12.88 & 58.12 & 48.22 & 45.14 & 44.94 & 45.02 & 42.14 & 41.74 \\
FedAVMIXUP~\cite{lee2020adversarial} & 95.80 & 62.56 & 74.82 & 45.78 & 27.20 & 14.10 & 12.10 & 70.62 & 46.88 & 41.68 & 40.98 & 40.80 & 32.96 & 32.48\\
FedTRADES~\cite{zhang2019theoretically} & 96.44 & 63.56 & 75.38 & 42.38 & 25.56 & 14.34 & 13.50 & 66.94 & 53.68 & 48.54 & 48.52 & 48.44 & 45.68 & 45.46\\
CalFAT~\cite{chen2022calfat} & 96.66 & 63.64 & 75.52 & 41.98 & 25.54 & 16.83 & 13.12 & 68.06 & 54.24 & 50.26 & 50.18 & 50.00 & 45.60 & 45.30\\
DBFAT~\cite{zhang2023delving} & 96.80 & 63.92 & 75.78 & 43.28 & 26.53 & 15.60 & 13.94 & 69.64 & 54.56 & 51.98 & 50.78 & 50.18 & 46.84 & 46.00\\
\midrule
FedBAT (ours) & \textbf{97.06} & \textbf{68.54} & \textbf{78.62} & \textbf{48.54} & \textbf{31.58} & \textbf{17.68} & \textbf{16.18} & \textbf{73.40} & \textbf{59.26} & \textbf{53.78} & \textbf{53.58} & \textbf{53.42} & \textbf{50.36} & \textbf{50.28}\\
 \bottomrule
\end{tabular}}
\end{table*}

\begin{table*}[t]
\centering
    \caption{Clean accuracy (\%) and robust accuracy (\%) comparison of FedBAT and other baselines on SVHN, and Office-Amazon under non-IID setting with Dir(0.5). The best results for federated defense methods are in \textbf{bold}.}
    \vspace{-0.2cm}
    \label{tab:SO_CA_RA_comparison}
\resizebox{0.99\textwidth}{!}{
\begin{tabular}{l|ccccccc|ccccccc}
\toprule
Dataset & \multicolumn{7}{c|}{SVHN} & \multicolumn{7}{c}{Office-Amazon} \\
\midrule
Methods & Clean & FGSM & BIM & PGD-40 & PGD-100 & Square & AA  & Clean & FGSM & BIM & PGD-40 & PGD-100 & Square & AA  \\
\midrule
FedAvg~\cite{mcmahan2017communication} & 84.90 & 11.14 & 0.28 & 0.16 & 0.12 & 1.90 & 0.06 & 60.72 & 40.00 & 39.06 & 38.95 & 37.95 & 37.70 & 37.29 \\
MixFAT~\cite{zizzo2020fat} & 64.58 & 27.66 & 17.82 & 16.54 & 13.24 & 16.96 & 13.40 & 58.65 & 45.83 & 45.00 & 44.68 & 44.09 & 44.27 & 43.64\\
FedPGD~\cite{madry2018towards} & 62.94 & 27.48 & 18.21 & 15.97 & 15.58 & 18.72 & 13.74 & 58.12 & 47.08 & 46.45 & 45.48 & 42.45 & 43.83 & 45.41\\
FedALP~\cite{kannan2018adversarial}  & 64.70 & 25.18 & 18.08 & 15.10 & 14.11 & 17.02 & 14.00 & 59.49 & 46.97 & 46.56 & 46.45 & 44.41 & 45.20 & 44.89\\
FedAVMIXUP~\cite{lee2020adversarial} & 63.77 & 27.08 & 17.45 & 15.18 & 12.45 & 17.73 & 12.56 & 58.77 & 46.06 & 44.15 & 43.10 & 42.87 & 43.70 & 43.85\\
FedTRADES~\cite{zhang2019theoretically}& 62.24 & 24.08 & 17.30 & 16.28 & 14.30 & 16.82 & 13.64 & 58.43 & 47.29 & 47.08 & 46.97 & 45.36 & 46.04 & 45.41 \\
CalFAT~\cite{chen2022calfat} & 62.96 & 26.30 & 15.16 & 14.12 & 13.04 & 18.00 & 12.82 & 58.02 & 46.87 & 46.45 & 45.48 & 44.45 & 45.62 & 45.37\\
DBFAT~\cite{zhang2023delving} & 63.88 & 27.52 & 18.78 & 16.32 & 14.76 & 17.74 & 13.39 & 57.02 & 47.60 & 46.56 & 46.66 & 47.17 & 45.52 & 44.79\\
\midrule
FedBAT (ours) & \textbf{66.66} & \textbf{29.20} & \textbf{20.62} & \textbf{20.12} & \textbf{19.86} & \textbf{19.88} & \textbf{15.54} & \textbf{60.00} & \textbf{50.41} & \textbf{49.27} & \textbf{47.27} & \textbf{46.93} & \textbf{48.54} & \textbf{47.70}\\
 \bottomrule
\end{tabular}}
\end{table*}

\section{Experiments}
\label{sec:experiments}
\subsection{Experimental Setup}
\textbf{Datasets.} We experiment on five widely-used benchmark datasets, including MNIST~\cite{lecun1998gradient}, Fashion-MNIST~\cite{xiao2017fashion}, SVHN~\cite{netzer2011reading}, Office-Amazon~\cite{gong2012geodesic}, and CIFAR-10~\cite{krizhevsky2009learning}, to validate the potential benefits of FedBAT towards enhance edge intelligence. \textbf{MNIST} is a widely-used dataset containing 70,000 grayscale images of handwritten digits (0-9), commonly used for image classification tasks. \textbf{Fashion-MNIST} consists of 70,000 grayscale images representing 10 different categories of clothing and accessories, offering a more challenging compared to MNIST. \textbf{SVHN} is a dataset of real-world images of house numbers from Google Street View, designed for digit classification tasks. \textbf{Office-Amazon} is a domain adaptation dataset that merges images from the Office-31 and Caltech-256 datasets, featuring 10 shared categories of everyday objects. \textbf{CIFAR-10} includes 60,000 color images categorized into 10 classes, such as airplanes, cars, and animals.

\textbf{Baselines.} To assess the robustness of our proposed method alongside existing approaches, we utilize six mainstream attack techniques: FGSM~\cite{goodfellow2014explaining}, BIM~\cite{kurakin2018adversarial}, PGD~\cite{madry2018towards}, CW~\cite{carlini2017towards}, Square~\cite{andriushchenko2020square}, and AA~\cite{croce2020reliable}. For adversarial defense methods, we incorporate four well-established techniques—PGD\_AT~\cite{madry2018towards}, ALP~\cite{kannan2018adversarial}, AVMIXUP~\cite{lee2020adversarial}, and TRADES~\cite{zhang2019theoretically}—within the FL framework, referring to them as FedPGD, FedALP, FedAVMIXUP, and FedTRADES, respectively. Additionally, we compare FedBAT with three other state-of-the-art federated defense methods: MixFAT~\cite{zizzo2020fat}, CalFAT~\cite{chen2022calfat}, and DBFAT~\cite{zhang2023delving}. For a more comprehensive evaluation, we also include comparisons with FedAvg, which represents standard FL training without AT. All baseline methods are evaluated with 5 clients by default.

\textbf{Implementaion Details.} 
Following~\cite{zizzo2020fat,qiao2024logit,chen2022calfat,zhang2023delving}, we adopt a multi-layer CNN architecture for all tasks. To simulate non-IID data challenges among clients, we use the Dirichlet distribution~\cite{yurochkin2019bayesian}, which enables control over the degree of data heterogeneity. The distribution is governed by the parameter $\gamma$: a smaller $\gamma$ results in greater data imbalance across clients, while a larger $\gamma$ leads to a more even distribution of the data. We set $\gamma$ to 0.5 by default. Note that since MNIST, Fashion-MNIST, SVHN, and CIFAR-10 contain a large number of samples, we randomly sample only 10\% of the data from these datasets to verify our proposal by default due to resource limitations. Additionally, to enhance the diversity of representations generated from the original images, we randomly apply an augmentation method, such as random crop, flip, scale, or rotation, to the original images during each local epoch. Following~\cite{goodfellow2014explaining,qiao2024logit}, we adopt the following AT settings: for MNIST, the perturbation bound is set to 0.3 with a step size of 0.01; for Fashion-MNIST, the perturbation bound is 32/255 with a step size of 8/255; and for SVHN, Office-Amazon, and CIFAR-10, the perturbation bound is 8/255 with a step size of 2/255. Additionally, we use the SGD optimizer with a local batch size of 64, 1 local epoch per global round, 100 global communication rounds, and a learning rate of 0.01. Note that for a fair comparison, all methods use the same network architecture and training parameters.

\subsection{Accuracy Comparison}
We implement FedBAT and all baseline methods using PyTorch and conduct preliminary comparisons of clean and robust accuracy (CA and RA) under non-IID settings. The results are shown in Tables~\ref{tab:M_F_CA_RA_comparison}, ~\ref{tab:SO_CA_RA_comparison}, and~\ref{tab:cifar10_CA_RA_comparison} with metrics averaged over the last 5 iterations. Overall, FedBAT consistently outperforms all baseline methods, achieving significant improvements in both clean and robust accuracy metrics.

\begin{table}[t]
\centering
\caption{Clean accuracy (\%) and robust accuracy (\%) comparison on CIFAR-10 with Dir(0.5). The best results for federated defense methods are in \textbf{bold}.}
\label{tab:cifar10_CA_RA_comparison}
\vspace{-0.2cm}
\resizebox{0.49\textwidth}{!}{
\begin{tabular}{l|cccccccc}
\toprule
Dataset & \multicolumn{6}{c}{CIFAR-10} \\
\midrule
Methods & Clean & FGSM & BIM & PGD-40 & PGD-100 & Square & AA  \\
\midrule
FedAvg~\cite{mcmahan2017communication} & 58.06 & 3.96 & 1.60 & 1.50 & 1.46 & 3.16 & 1.20  \\
MixFAT~\cite{zizzo2020fat} & 44.82 & 23.10 & 20.60 & 20.54 & 20.40 & 19.80 & 17.14  \\
FedPGD~\cite{madry2018towards} & 43.28 & 25.56 & 24.32 & 24.34 & 24.28 & 21.60 & 20.24  \\
FedALP~\cite{kannan2018adversarial} & 42.00 & 25.52 & 25.18 & 24.38 & 24.10 & 21.55 & 20.28 \\
FedAVMIXUP~\cite{lee2020adversarial} & 44.02 & 23.53 & 22.88 & 24.03 & 23.84 & 20.66 & 20.12 \\
FedTRADES~\cite{zhang2019theoretically} & 44.48 & 25.66 & 24.36 & 24.28 & 24.08 & 21.48 & 20.16  \\
CalFAT~\cite{chen2022calfat} & 43.94 & 25.16 & 23.94 & 23.96 & 23.77 & 21.50 & 20.28  \\
DBFAT~\cite{zhang2023delving} & 44.34 & 25.62 & 24.26 & 24.28 & 24.06 & 21.36 & 19.84  \\
\midrule
FedBAT (ours) & \textbf{45.98} & \textbf{26.00} & \textbf{25.64} & \textbf{24.72} & \textbf{24.64} & \textbf{22.60} & \textbf{20.70}  \\
\bottomrule
\end{tabular}}
\vspace{-5px}
\end{table}

Table~\ref{tab:M_F_CA_RA_comparison} presents the results for MNIST and Fashion-MNIST tasks. The `clean' column shows the accuracy of adversarially trained models evaluated on clean samples, while FGSM, BIM, PGD-40, PGD-100, Square, and AA represent adversarial attacks used to assess robustness on adversarial samples. Among these, AA is widely regarded as the most powerful attack algorithm, as it incorporates multiple advanced attack strategies. Taking the Fashion-MNIST task as an example, several observations can be made. First, the performance of all methods, including ours, declines under adversarial attacks, with the lowest accuracy observed under AA attacks. This is particularly evident in the MNIST task, where only around 13\% accuracy is achieved in the AA attack, compared to over 60\% accuracy in the FGSM attack. This confirms that AA is the strongest attack algorithm and poses significant challenges to federated models. Second, the conventional federated learning algorithm, FedAvg, is vulnerable to adversarial attacks, as its robustness under AA attacks drops significantly from 79.18\% to 10.34\%. This highlights our motivation to improve the robustness of the model in the federated setting. Third, with our proposed defense strategy, FedBAT outperforms all other methods across both clean and robust accuracy metrics. Specifically, compared to the baseline MixFAT, FedBAT demonstrates a 2.18\% increase in clean accuracy and a notable 6.71\% improvement in robust accuracy, with the robust accuracy averaged over the six attack types. A similar trend is observed in Tables~\ref{tab:SO_CA_RA_comparison} and~\ref{tab:cifar10_CA_RA_comparison} for more challenging tasks, with the former reporting results for SVHN and Office-Amazon, and the latter for CIFAR-10. These results across diverse tasks further confirm the three observations above and emphasize the effectiveness of our proposed FedBAT in enhancing robustness while maintaining high clean accuracy.

\begin{figure}[t]
\centering
\includegraphics[width=0.40\textwidth]{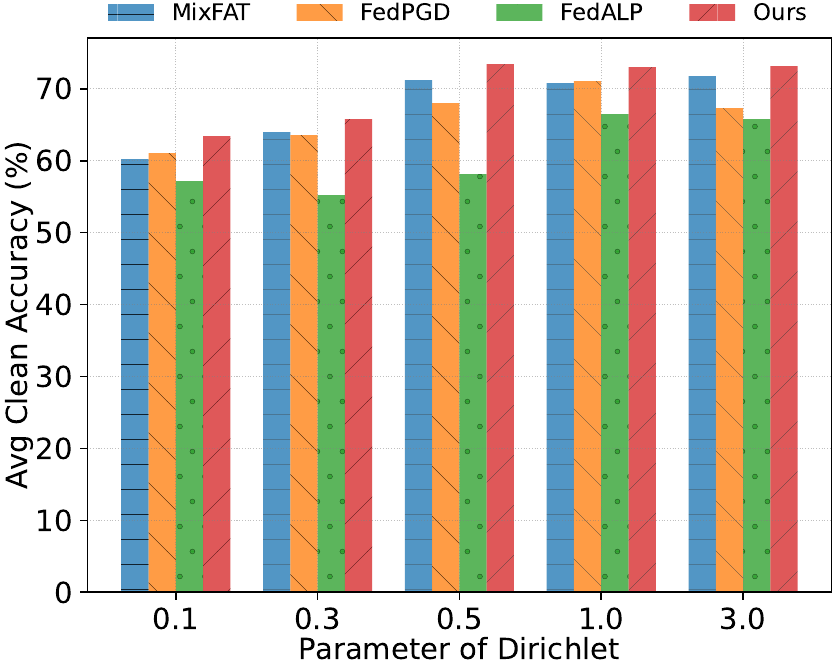}
% \vspace{-0.2cm}
\caption{Comparison of the average clean accuracy (CA) on Fashion-MNIST under different levels of data heterogeneity. The clean accuracy is reported for the adversarially trained model but evaluated on clean samples. A lower Dirichlet parameter value indicates higher heterogeneity.}
\label{fig:diff_dirichlet_fashionmnist_nonIID}
\vspace{-5px}
\end{figure}

\subsection{Robustness Comparison}
\begin{figure}[t]
\centering
\includegraphics[width=0.40\textwidth]{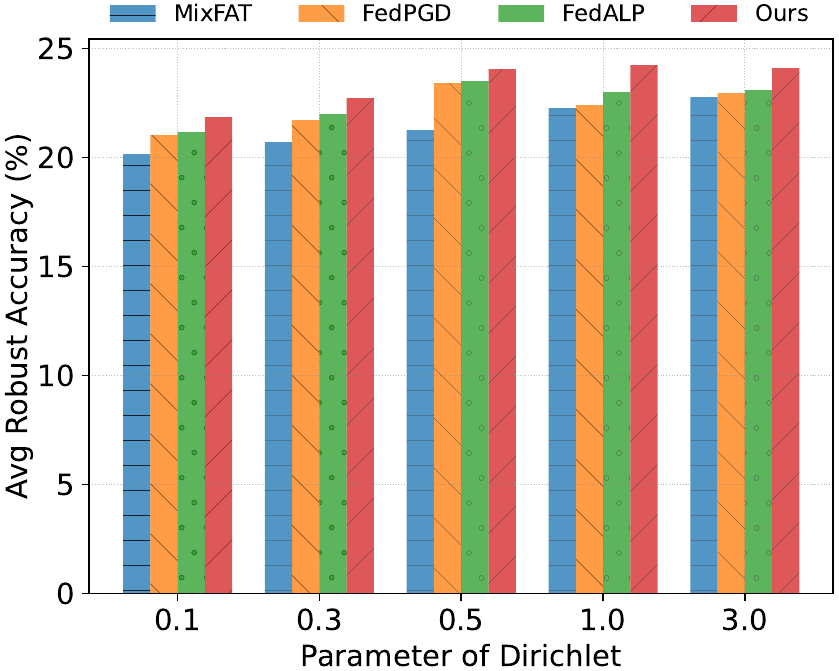}
% \vspace{-0.2cm}
\caption{Comparison of the average robust accuracy (RA) on CIFAR-10 under different levels of data heterogeneity. The robust accuracy is averaged across six attack methods: FGSM, BIM, PGD-40, PGD-100, Square, and AA attacks. A lower Dirichlet parameter value indicates higher heterogeneity.}
\label{fig:diff_dirichlet_cifar10_nonIID}
\vspace{-5px}
\end{figure}

\textbf{Different Heterogeneities.} As discussed in the previous section, the non-IID data issue presents significant challenges in federated adversarial settings. To ensure successful practical deployment in the future, it is critical to evaluate the robustness of algorithms under varying degrees of data heterogeneity. Figure~\ref{fig:diff_dirichlet_fashionmnist_nonIID} shows the clean accuracy comparison, and Figure~\ref{fig:diff_dirichlet_cifar10_nonIID} reports the robustness comparison. Again, clean accuracy (CA) is measured on adversarially trained models using clean samples, while robust accuracy (RA) is averaged across six attack methods—FGSM, BIM, PGD-40, PGD-100, Square, and AA—for comprehensive evaluation.

Figure~\ref{fig:diff_dirichlet_fashionmnist_nonIID} shows the clean accuracy for Fashion-MNIST across a wide range of heterogeneity settings, including 0.1, 0.3, 0.5, 1.0, and 3.0. An overall observation from the figure is that FedBAT (ours) consistently outperforms other baselines, highlighting the effectiveness of our proposals in handling different levels of heterogeneity. For example, with the Dirichlet parameter set to 0.1, which creates a significantly imbalanced data distribution among clients, the clean accuracy of FedBAT still surpasses that of the second-best method, FedPGD, by around 2\%. For adversarial robustness comparison, Figure~\ref{fig:diff_dirichlet_cifar10_nonIID} illustrates the results for CIFAR-10 under the same range of data heterogeneity. A similar observation can be made that FedBAT consistently outperforms other baselines in terms of robustness across different Dirichlet parameters in our experimental settings. For example, as the Dirichlet parameter increases from 0.1 to 3.0, the robustness of FedBAT rises from 21.85\% to 24.10\%, while that of FedPGD increases from 21.00\% to 22.93\%. All of these observations confirm that our proposal is robust against various heterogeneities, achieving higher robust accuracy while maintaining clean accuracy compared to these baselines.

\begin{table*}[t]
\centering
    \caption{Scalability comparison (\%) of FedBAT and other baselines on MNIST and Fashion-MNIST with a large number of clients, where the participation rate is set to 0.1. The best results for federated defense methods are in \textbf{bold}.}
    \label{tab:M_F_scala_comparison}
    \vspace{-0.2cm}
\resizebox{0.99\textwidth}{!}{
\begin{tabular}{l|ccccccc|ccccccc}
\toprule
Dataset & \multicolumn{7}{c|}{MNIST} & \multicolumn{7}{c}{Fashion-MNIST} \\
\midrule
Methods & Clean & FGSM & BIM & PGD-40 & PGD-100 & Square & AA & Clean & FGSM & BIM & PGD-40 & PGD-100 & Square & AA \\
\midrule
FedAvg~\cite{mcmahan2017communication} & 92.88 & 3.64 & 16.24 & 1.36 & 0.72 & 0.52 & 0.28 & 68.56 & 22.08 & 14.96 & 14.68 & 14.52 & 13.80 & 13.28 \\
MixFAT~\cite{zizzo2020fat} & 82.34 & 25.94 & 43.34 & 14.54 & 9.39 & 7.89 & 4.97 & 52.78 & 37.42 & 34.89 & 36.90 & 36.51 & 33.24 & 31.91 \\
FedPGD~\cite{madry2018towards} & 81.04 & 26.28 & 45.12 & 14.96 & 9.12 & 8.36 & 5.04 & 51.28 & 40.08 & 37.76 & 37.64 & 37.16 & 34.20 & 33.28 \\
FedALP~\cite{kannan2018adversarial} & 79.12 & 27.52 & 45.56 & 14.16 & 10.64 & 8.16 & 5.09 & 50.76 & 36.36 & 33.84 & 34.04 & 34.02 & 24.64 & 24.24\\
FedAVMIXUP~\cite{lee2020adversarial} & 80.91 & 25.97 & 43.01 & 16.22 & 10.33 & 8.68 & 4.93 & 49.93 & 39.33 & 37.66 & 35.82 & 37.34 & 33.46 & 33.34\\
FedTRADES~\cite{zhang2019theoretically} & 81.32 & 26.56 & 45.68 & 15.20 & 9.48 & 8.84 & 5.08 & 51.36 & 40.20 & 38.12 & 38.08 & 37.08 & 33.92 & 33.08\\
CalFAT~\cite{chen2022calfat} & 80.92 & 26.48 & 44.96 & 15.08 & 9.28 & 8.52 & 5.08 & 51.40 & 40.00 & 37.76 & 37.56 & 37.52 & 33.52 & 32.68\\
DBFAT~\cite{zhang2023delving} & 80.80 & 26.04 & 45.04 & 15.20 & 9.44 & 8.40 & 4.72 & 52.24 & 40.96 & 38.80 & 38.88 & 38.72 & 33.84 & 33.00\\
\midrule
FedBAT (ours) & \textbf{83.20} & \textbf{31.88} & \textbf{47.24} & \textbf{16.60} & \textbf{11.04} & \textbf{9.00} & \textbf{5.52} & \textbf{55.24} & \textbf{42.68} & \textbf{41.16} & \textbf{41.12} & \textbf{41.02} & \textbf{34.80} & \textbf{34.12}\\
\bottomrule
\end{tabular}}
\end{table*}

\begin{table}[!htbp]
\centering
\caption{Scalability comparison (\%) of FedBAT and other baselines on Cifar-10 with a large number of clients, where the participation rate is set to 0.1. The best results for federated defense methods are in \textbf{bold}.}
\label{tab:cifar10_scala_comparison}
\vspace{-0.2cm}
\resizebox{0.479\textwidth}{!}{
\begin{tabular}{l|cccccccc}
\toprule
Dataset & \multicolumn{6}{c}{Cifar-10} \\
\midrule
Methods & Clean & FGSM & BIM & PGD-40 & PGD-100 & Square & AA  \\
\midrule
FedAvg~\cite{mcmahan2017communication} & 37.24 & 12.68 & 11.08 & 11.04 & 10.98 & 10.64 & 8.84 \\
MixFAT~\cite{zizzo2020fat} & 24.69 & 17.47 & 16.42 & 17.07 & 16.65 & 15.28 & 14.77 \\
FedPGD~\cite{madry2018towards} & 24.40 & 18.24 & 16.93 & 17.32 & 17.56 & 15.56 & 15.60 \\
FedALP~\cite{kannan2018adversarial} & 23.88 & 18.08 & 17.64 & 17.60 & 17.52 & 15.16 & 15.56 \\
FedAVMIXUP~\cite{lee2020adversarial} & 24.21 & 17.83 & 16.94 & 17.00 & 16.88 & 16.03 & 14.98 \\
FedTRADES~\cite{zhang2019theoretically} & 24.16 & 18.28 & 16.95 & 17.84 & 16.94 & 15.60 & 15.48 \\
CalFAT~\cite{chen2022calfat} & 24.36 & 18.20 & 17.64 & 17.72 & 17.59 & 15.48 & 14.83 \\
DBFAT~\cite{zhang2023delving} & 24.84 & 18.26 & 17.58 & \textbf{18.03} & 17.62 & 15.68 & 15.40 \\
\midrule
FedBAT (ours) & \textbf{25.32} & \textbf{18.44} & \textbf{17.80} & 17.92 & \textbf{17.76} & \textbf{16.56} & \textbf{15.84} \\
\bottomrule
\end{tabular}}
\vspace{-0.35cm}
\end{table}

\begin{figure}[t]
\centering
\includegraphics[width=0.40\textwidth]{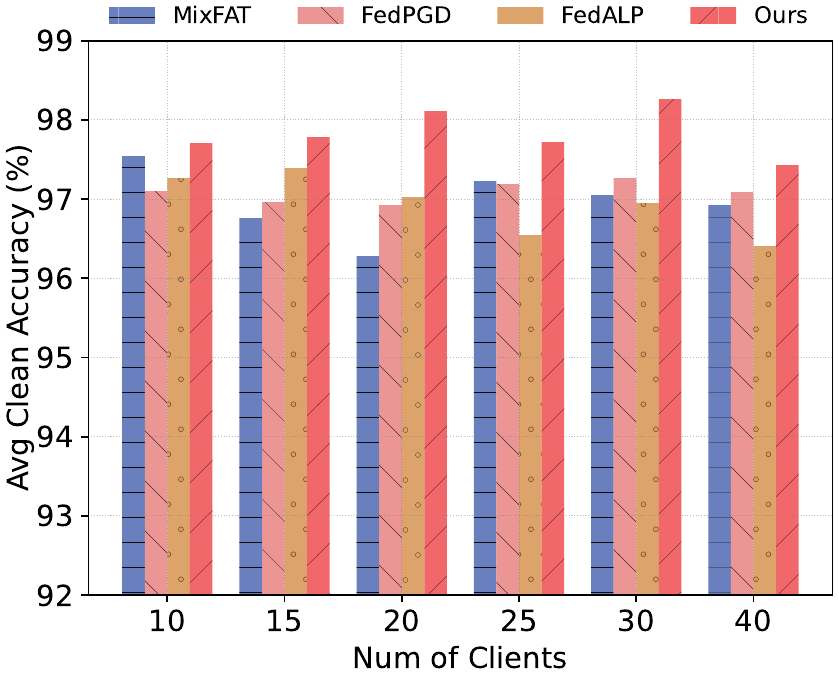}
% \vspace{-0.2cm}
\caption{Comparison of the average clean accuracy (CA) on MNIST under different numbers of clients. The clean accuracy is reported for the adversarially trained model but evaluated on clean samples.}
\vspace{-15px}
\label{fig:diff_num_mnist}
\end{figure}

\begin{figure}[t]
\centering
\includegraphics[width=0.40\textwidth]{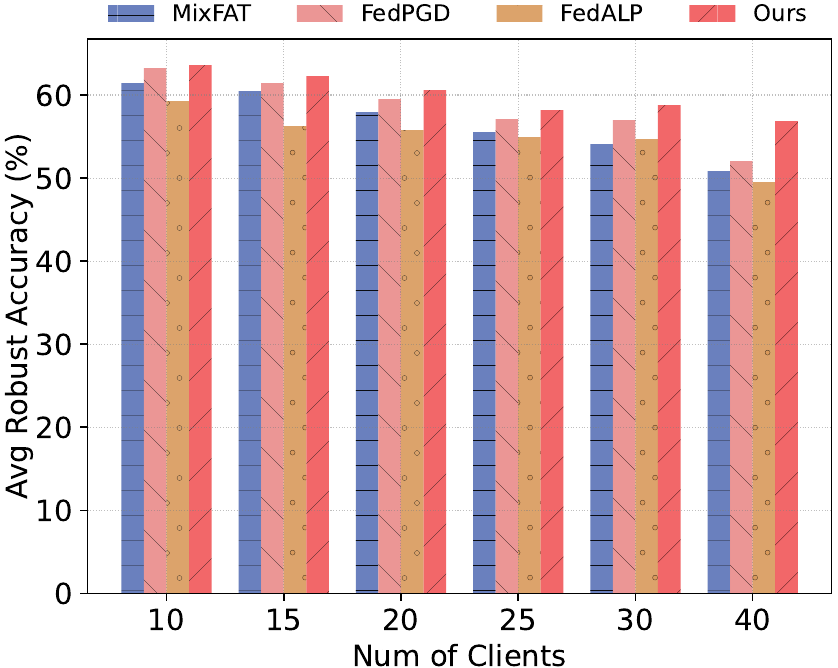}
% \vspace{-0.2cm}
\caption{Comparison of the average robust accuracy (RA) on Fashion-MNIST under different numbers of clients. The robust accuracy is averaged across six attack methods: FGSM, BIM, PGD-40, PGD-100, Square, and AA attacks.}
\label{fig:diff_num_fashion_mnist}
\vspace{-5px}
\end{figure}

\textbf{Different Numbers of Clients.} As the number of clients increases, the sample distribution for each client decreases, creating additional challenges for the federated training process. Therefore, for completeness, we further evaluate the robustness of our proposed method under these conditions. However, since using only 10\% of the training samples can result in some clients not receiving any samples, we include all samples in the training process for this robustness evaluation. The results are reported in Figures~\ref{fig:diff_num_mnist} and~\ref{fig:diff_num_fashion_mnist}.

Specifically, Figure~\ref{fig:diff_num_mnist} presents the clean accuracy for MNIST, while Figure~\ref{fig:diff_num_fashion_mnist} illustrates the robust accuracy for Fashion-MNIST. Both figures examine a range of client numbers, specifically 10, 15, 20, 25, 30, and 40. All methods are evaluated using a Dirichlet distribution with a parameter of 0.5. From the results, several observations can be made. First, our proposal consistently outperforms several baselines in terms of both accuracy and robustness metrics, as indicated by the values represented by the red columns in both figures. Second, as the number of clients increases, each client receives fewer samples, which presents a greater challenge for federated algorithms. This is evident from the decrease in robust accuracy in Figure~\ref{fig:diff_num_fashion_mnist}. However, our proposed method exhibits greater robustness than the baseline across various client number scenarios, effectively demonstrating its robustness to such variations. For instance, in the Fashion-MNIST task with 40 clients, FedBAT achieves an overall robustness gain of 9.34\% compared to FedPGD.

\subsection{Scalability Comparison}
To demonstrate the effectiveness and scalability of our proposal in a large-scale setting involving a substantial number of clients, we conduct a scalability comparison in which only a subset of clients participates in the federated training process during each global round. Specifically, to ensure that all clients receive samples, we utilize the entire dataset and partition it into 100 clients using the default Dirichlet parameter value, setting the participation rate to 0.1 for each communication round. As a result, in each global communication round, 10 clients participate in the federated training process. The results are presented in Table~\ref{tab:M_F_scala_comparison} for the MNIST and Fashion-MNIST tasks, and in Table~\ref{tab:cifar10_scala_comparison} for the CIFAR-10 task. From the results, we can observe a general trend that our proposal remains consistently effective in large-scale settings and outperforms other baselines on different datasets in most cases, underscoring its potential for practical applications. For example, in terms of clean accuracy, FedBAT outperforms the baseline FedPGD by 7.72\% on Fashion-MNIST, and even on the more challenging CIFAR-10 dataset, our method still surpasses FedPGD by 3.77\%. Regarding robustness, FedBAT exceeds the baseline FedPGD by 6.92\% on the AA attack metric for Fashion-MNIST, and on the CIFAR-10 task, our method again outperforms FedPGD by 1.53\% on the same metric. Note that while our approach slightly lags behind DBFAT on CIFAR-10 under the PGD-40 metric, it achieves a 1.69\% improvement in average robust accuracy compared to DBFAT.

\begin{figure*}[!htbp]
\centering
\subfloat[MNIST]{\includegraphics[width = 0.21\textwidth]{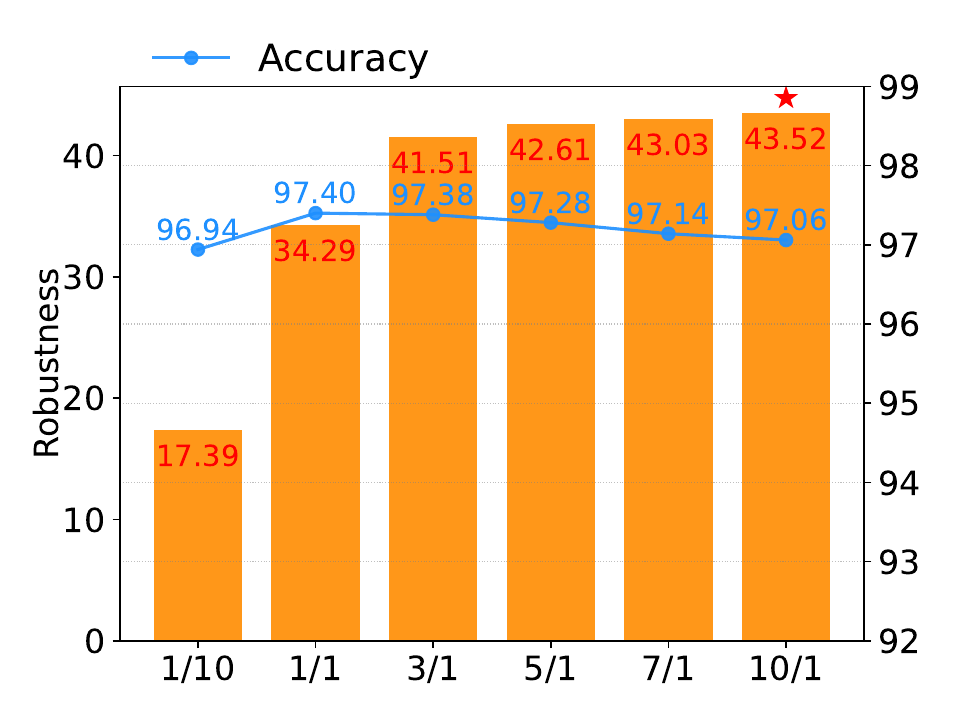}}
% \hfill
\subfloat[Fashion-MNIST]{\includegraphics[width = 0.20\textwidth]{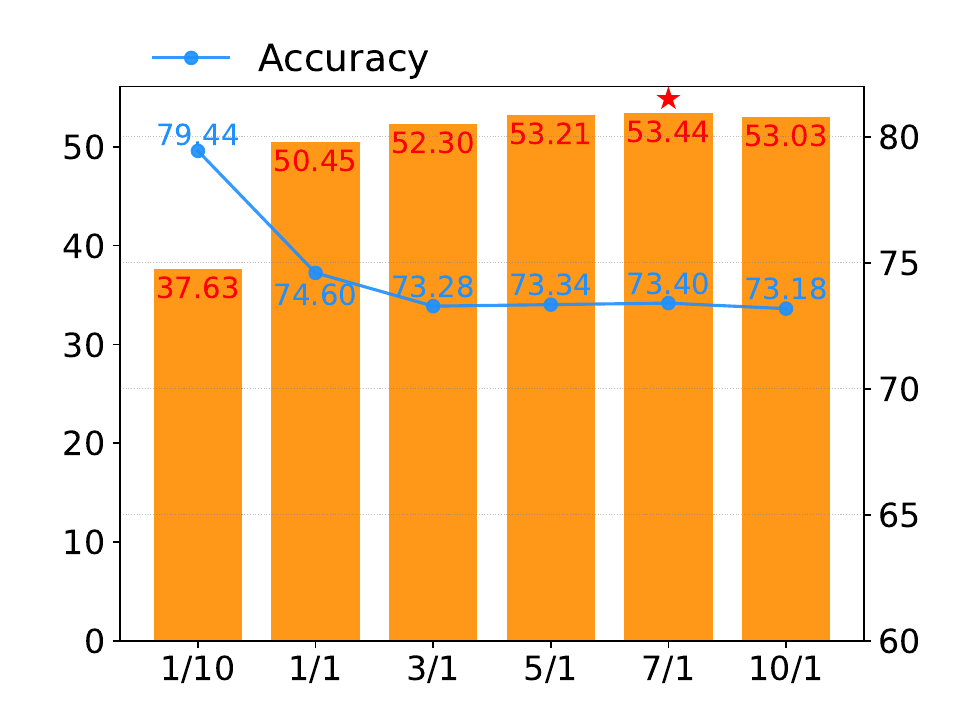}}
% \hfill
\subfloat[Cifar-10]{\includegraphics[width = 0.20\textwidth]{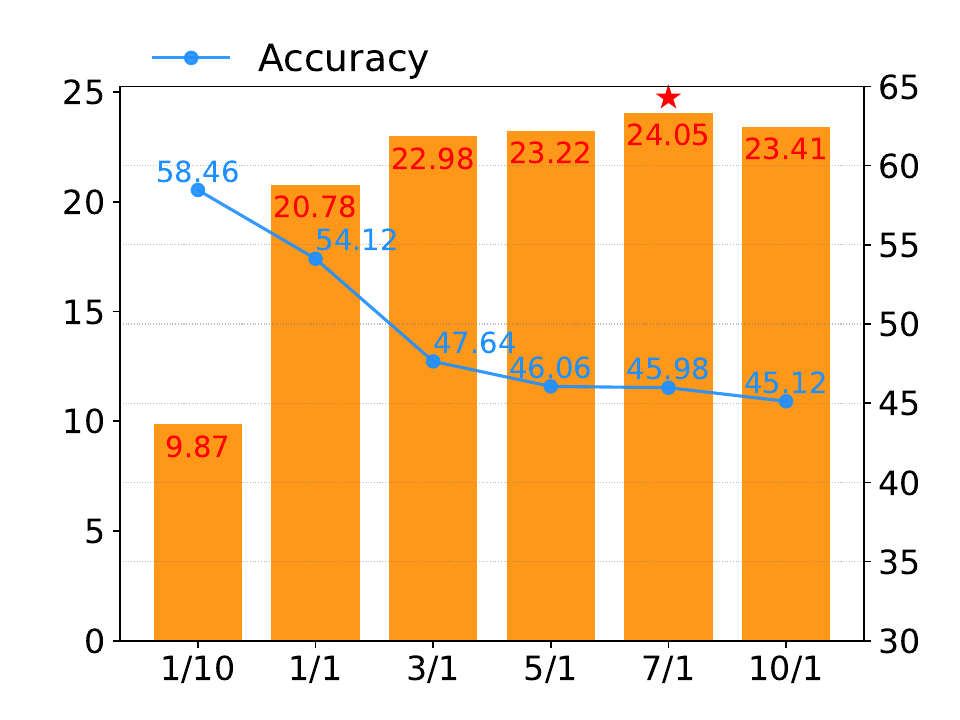}} 
% \hfill
\subfloat[SVHN]{\includegraphics[width = 0.20\textwidth]{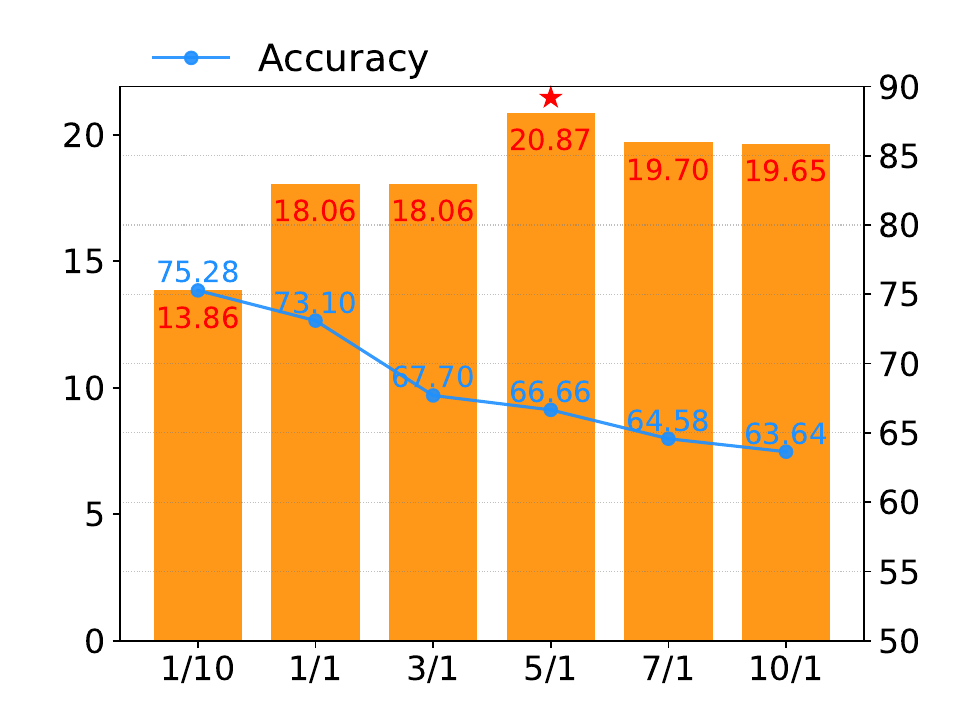}} 
% \hfill
\subfloat[Office-Amazon]{\includegraphics[width = 0.20\textwidth]{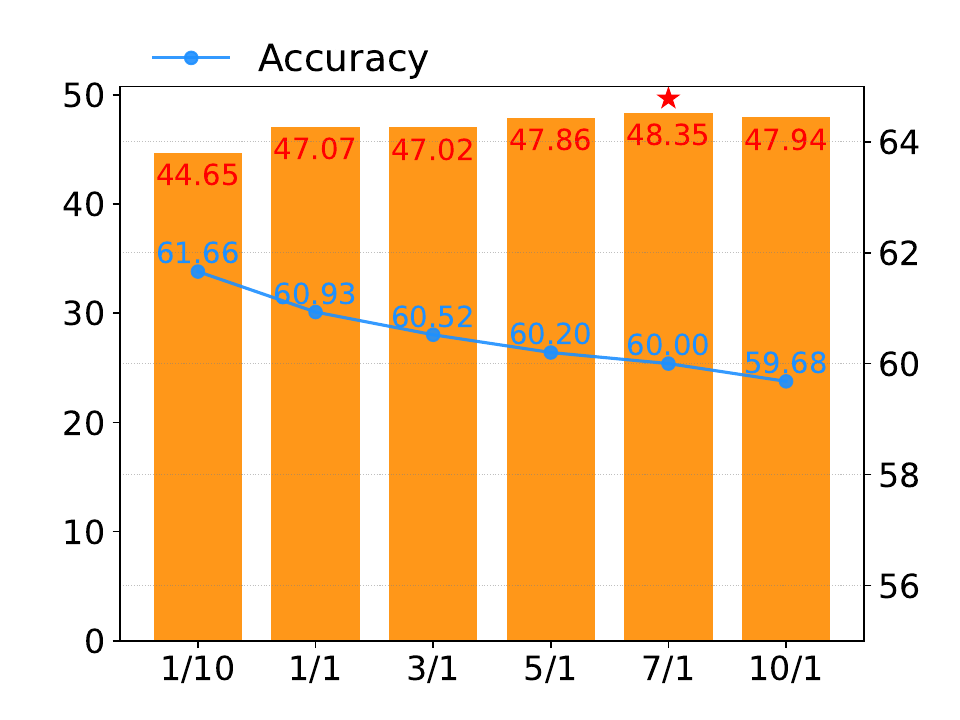}} 
\caption{Tradeoff between accuracy and robustness across various tasks under different values of $\rho$. The X-axis represents the ratio of robustness to accuracy, defined as $\rho = \frac{\lambda}{1-\lambda}$. The left Y-axis shows robustness, while the right Y-axis indicates accuracy. The selected parameter for each task is highlighted with \textcolor{red}{\FiveStar}. From left to right, the selected values of $\rho$ are 10.0/1.0, 7.0/1.0, 7.0/1.0, 5.0/1.0, and 7.0/1.0, respectively.}
\label{fig:choose_lambda}
\end{figure*}

\subsection{Ablation Study}
We perform ablation studies to analyze the effectiveness of each module in our proposed framework to the overall performance. Table~\ref{tab:ablation_comparison} presents the results across various tasks, from which several key observations can be drawn. First, when neither the hybrid-AT nor the augmentation-invariant self-adversarial distillation strategies are applied during local training, performance declines across these tasks. For instance, on the Office-Amazon task, the clean accuracy drops from 60.00\% to 58.12\%, and the robust accuracy decreases from 48.35\% to 45.11\%. Second, we observe that incorporating the hybrid-AT strategy, along with a carefully chosen parameter for balancing accuracy and robustness, leads to noticeable improvements across all datasets. For example, on the Fashion-MNIST task, applying hybrid-AT alone boosts clean accuracy from 68.00\% to 71.80\% and robust accuracy from 48.86\% to 49.60\%. This indicates that hybrid-AT can enhance both clean and robust performance. Third, when the ASD strategy without augmentation (Config2) is added, further gains in performance are observed. On the Office-Amazon task, clean accuracy increases from 58.42\% to 59.68\%, and robust accuracy rises from 46.38\% to 47.83\%. This further suggests that the ASD strategy can improve performance by effectively guiding each local training. Finally, the best results are achieved when both the hybrid-AT and the full ASD strategy with augmentation (Config2 and Config3) are utilized. In every task, this configuration yields the highest clean and robust accuracy, demonstrating the overall effectiveness of combining both strategies with augmentation.

\begin{table}[t]
\centering
\caption{Ablation study of different configurations within the proposed framework across various datasets. `Config1' refers to hybrid-AT, `Config2' denotes the ASD strategy without augmentation, and `Config3' represents the ASD strategy with augmentation.}
\vspace{-0.2cm}
\label{tab:ablation_comparison}
\resizebox{0.99\linewidth}{!}{
\begin{tabular}{c|ccc|cccc}
\toprule
Dataset & Config1 & Config2 & Config3  & CA & $\triangle$  & RA & $\triangle$  \\
\midrule
MNIST & \xmark & \xmark & \xmark & 96.44 & - & 37.71 & -\\
& \cmark & \xmark & \xmark  & 96.64 & +0.20 & 38.02 & +0.31\\
& \cmark & \cmark & \xmark & 96.73 & +0.29 & 40.80 & +3.09 \\
& \cmark & \cmark & \cmark & \textbf{97.06} & +0.62 & \textbf{43.52}  & +5.81\\
\midrule
SVHN & \xmark & \xmark & \xmark & 62.94 & - & 18.28  & -\\
& \cmark & \xmark & \xmark & 63.72 & +0.78 & 18.67 & +0.39  \\
& \cmark & \cmark & \xmark & 65.00 & +2.06 & 19.21  & +0.93\\
& \cmark & \cmark & \cmark & \textbf{66.66}  & +3.72 & \textbf{20.87} & +2.59\\
\midrule
Cifar-10 & \xmark & \xmark & \xmark & 43.28 & - & 23.39 & - \\
& \cmark & \xmark & \xmark  & 44.84 & +1.56 & 23.53 & +0.14 \\
& \cmark & \cmark & \xmark & 45.18 & +1.90 & 23.60 & +0.21 \\
& \cmark & \cmark & \cmark & \textbf{45.98} & +2.70 & \textbf{24.05} & +0.66 \\
\midrule
Fashion-MNIST & \xmark & \xmark & \xmark & 68.00 & - & 48.86 & - \\
& \cmark & \xmark & \xmark  & 71.80 & +3.80 & 49.60 & +0.74 \\
& \cmark & \cmark & \xmark & 72.88 & +4.88 & 53.00 & +4.14\\
& \cmark & \cmark & \cmark & \textbf{73.40} & +5.40 & \textbf{53.44} & +4.58\\
\midrule
Office-Amazon & \xmark & \xmark & \xmark & 58.12 & - & 45.11 & - \\
& \cmark & \xmark & \xmark  & 58.42 & +0.30 & 46.38 & +1.27 \\
& \cmark & \cmark & \xmark & 59.68 & +1.56 & 47.83 & +2.72 \\
& \cmark & \cmark & \cmark & \textbf{60.00} & +1.88 & \textbf{48.35} & +3.24 \\
\bottomrule
\end{tabular}}
% \vspace{-0.3cm}
\end{table}

\subsection{Tradeoff Between Accuracy and Robustness}
\label{apex_parameter_selection}
In this section, we explore the most suitable hyper-parameter $\lambda$ in (\ref{fl_hybrid_at}) for balancing accuracy and robustness under different tasks. We conduct experiments on various tasks, presenting the relationship between accuracy and robustness as a function of $\rho = \frac{\lambda}{1-\lambda}$, which represents the ratio of the adversarial loss to the clean loss. The corresponding accuracy and robustness are reported in Figure~\ref{fig:choose_lambda}. It can be observed that as the ratio $\rho$ increases, accuracy decreases while robustness improves, which aligns with our intuition that the model shifts its focus toward robustness at the expense of clean accuracy. Furthermore, we find that both accuracy and robustness tend to reach a plateau after $\rho = 1.0$, and when $\rho$ exceeds 1.0, they fluctuate only slightly. This suggests that once accuracy and robustness reach a stable state, our method might become relatively insensitive to the hyperparameter. Nonetheless, since the objective of this paper is to improve the robustness of the global model while preserving relatively high accuracy, the most suitable parameter for each task is highlighted with a red star in the figure. Specifically, the selected values of $\rho$ are 10.0/1.0, 7.0/1.0, 7.0/1.0, 5.0/1.0, and 7.0/1.0, corresponding to MNIST, Fashion-MNIST, Cifar-10, SVHN, and Office-Amazon, respectively.

\subsection{Feature Visualization and Discussion}
\begin{figure}[t]
\centering
\subfloat[FedPGD (clean feature)]{\includegraphics[width = 0.23\textwidth]{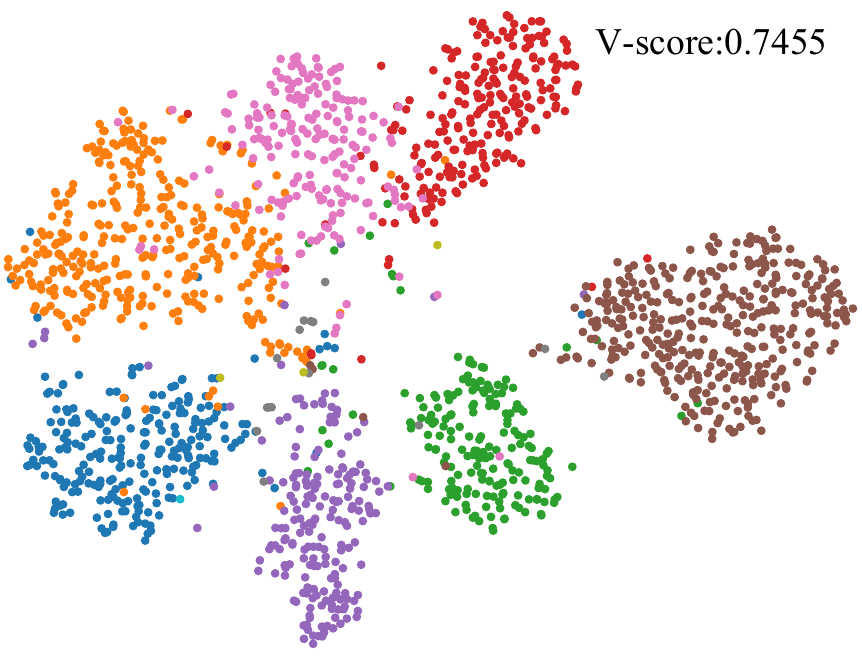}}
\hfill
\subfloat[FedPGD (adversarial feature)]{\includegraphics[width = 0.23\textwidth]{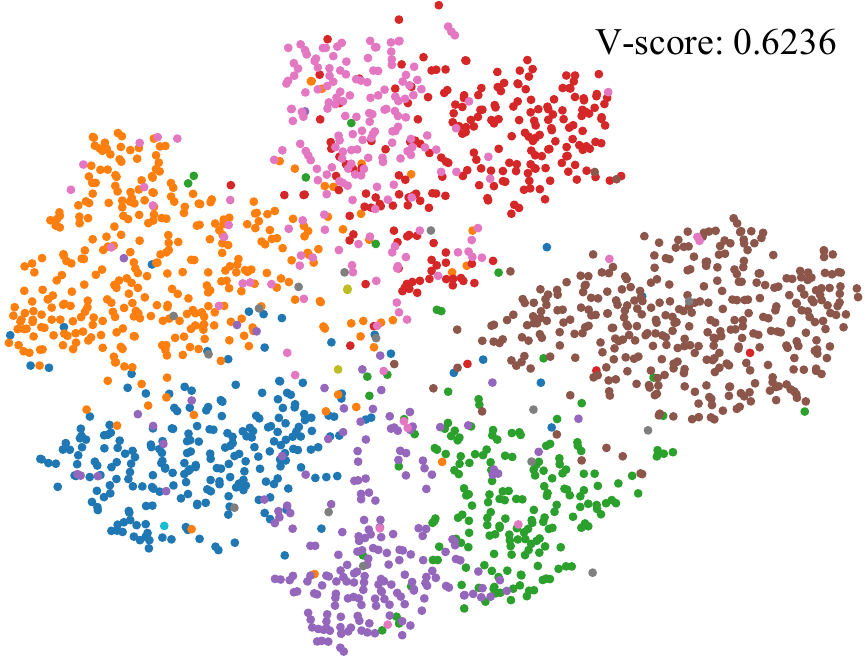}}
\hfill
\vspace{0.3cm}
\subfloat[FedBAT (clean feature)]{\includegraphics[width = 0.23\textwidth]{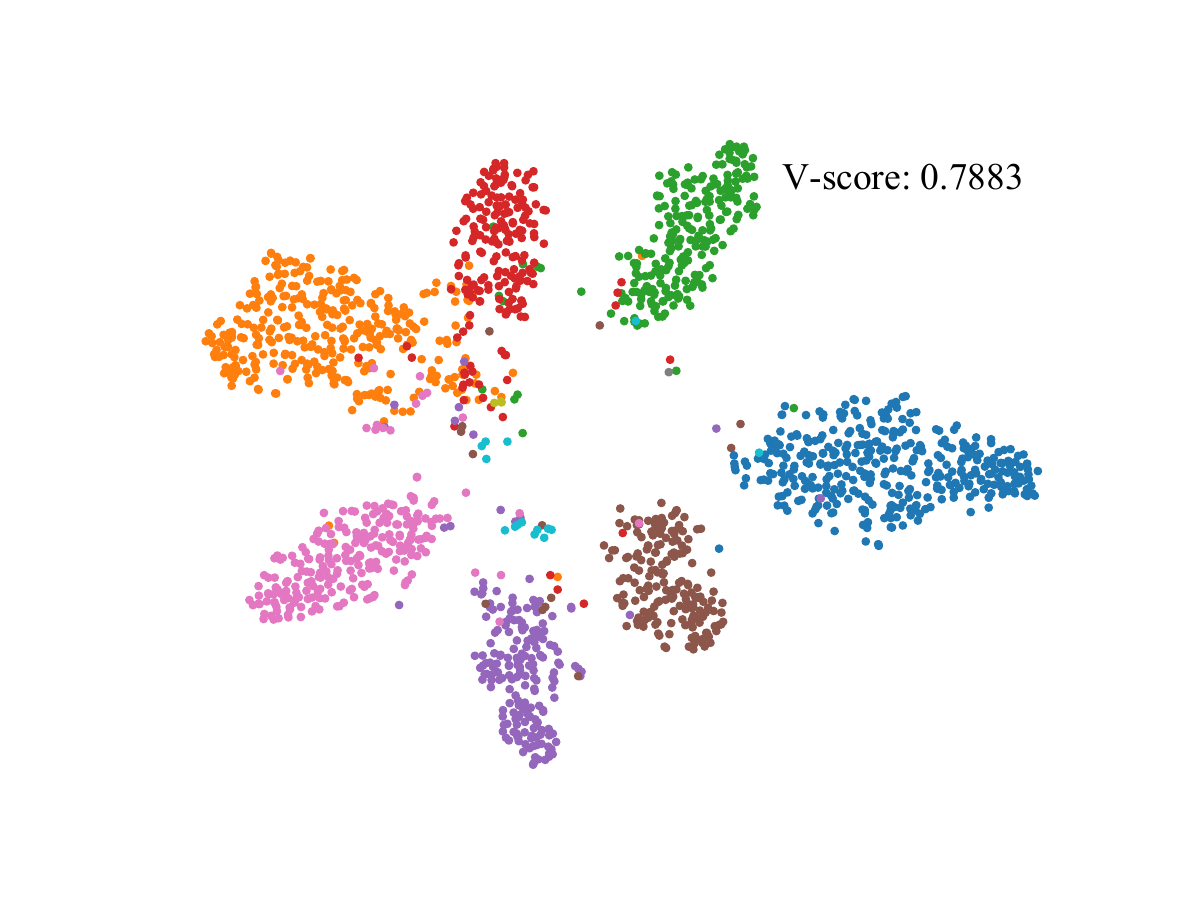}}
\hfill
\subfloat[FedBAT (adversarial feature)]{\includegraphics[width = 0.23\textwidth]{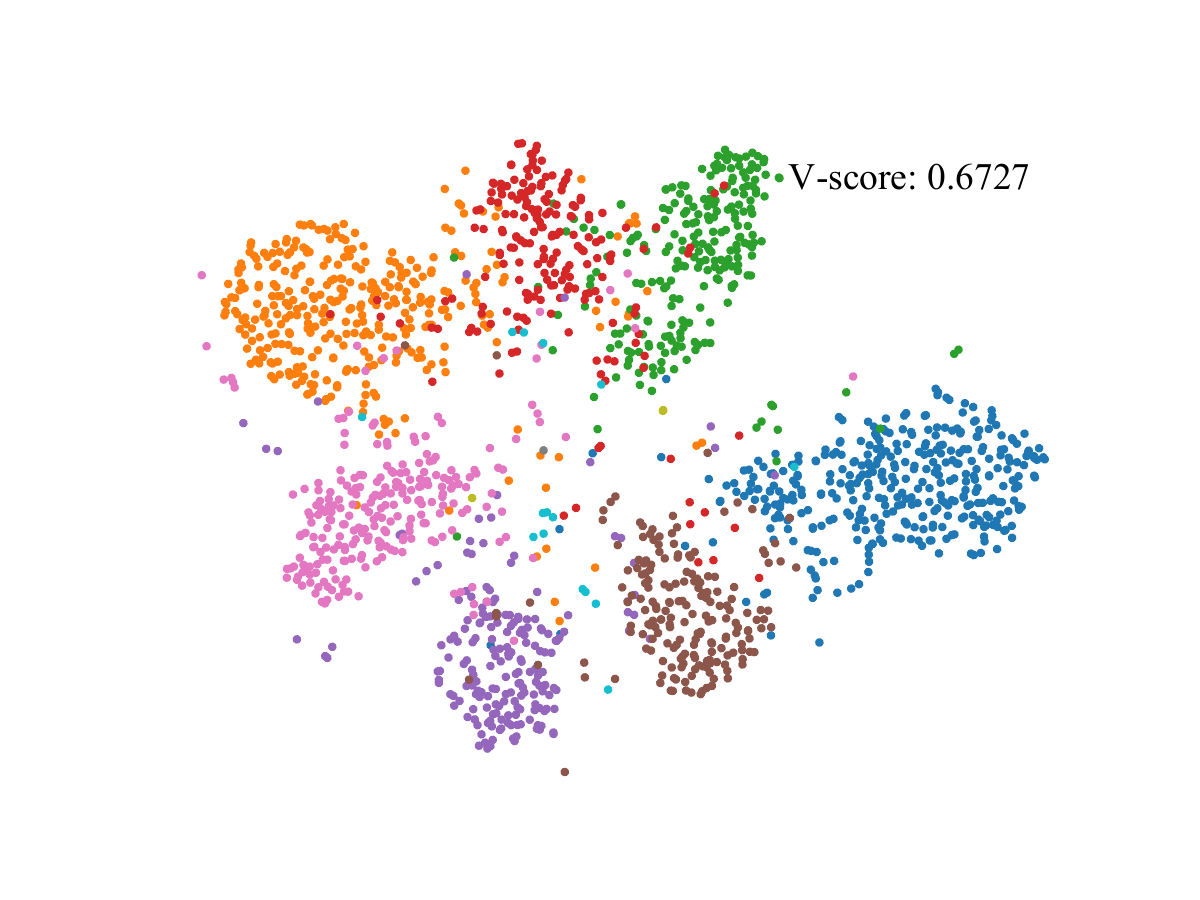}}
\hfill
\caption{Feature visualization extracted from the same client for FedPGD and FedBAT under the MNIST task. In either the first row or the second row, we observe that clean features promote tighter clustering in each class (higher V-score) than adversarial features. In contrast, both in clean feature and adversarial feature visualization, FedBAT achieves tighter clustering in each class compared to FedPGD.}
\label{fig:visualization}
\vspace{-11px}
\end{figure}

To provide a clear insight into our proposal, we present t-SNE visualizations of our method and the baseline in Figure~\ref{fig:visualization}. The first row shows both clean and adversarial feature embeddings of FedPGD, where clean features are derived from clean samples and adversarial features from adversarial samples. As depicted, the clean feature embeddings exhibit tighter clustering, while the adversarial feature embeddings are more dispersed. This observation is further supported by the V-score~\cite{rosenberg2007v} clustering metric, where higher values indicate better clustering. A similar trend is seen in the second row, which shows the results for FedBAT. This observation may explain why robust accuracy is lower than clean accuracy, as adversarial features have less distinct inter-class boundaries and more dispersed intra-class distributions compared to clean features. Interestingly, we also observe that our method achieves tighter clustering in both clean and adversarial feature visualizations compared to the baselines. This suggests the effectiveness of our approach, where hybrid-AT enhances robustness while maintaining higher accuracy, and self-adversarial distillation encourages the model to inherit a more discriminative embedding from the global model.

\section{Conclusion}
\label{sec:conclusion}
In this paper, we have proposed a new robust federated learning framework aimed at addressing both adversarial attacks and non-IID data challenges. We begin by pointing out the negative impact of adversarial attacks on performance and the limitations of vanilla robust federated algorithms like FedPGD, which overlook clean accuracy and the discrepancies between local and global models. Building on these observations, we first propose a hybrid adversarial training strategy that treats adversarial training as a form of data augmentation to improve robustness while maintaining clean accuracy. Second, we propose a novel augmentation-invariant self-adversarial distillation strategy to ensure consistency between the updates of local models and the global model, thereby enhancing the model's generalization ability. Experimental results across multiple tasks demonstrate that our proposed method consistently outperforms various baselines, achieving a 5.40\% improvement in clean accuracy and a 4.58\% boost in robust accuracy over FedPGD on Fashion-MNIST, with corresponding gains of 3.72\% and 2.59\% on SVHN. Similar improvements are observed across other datasets, further verifying the effectiveness and robustness of our approach.

\bibliographystyle{ieeetr}
\bibliography{bib_global}

\end{document}